\pdfoutput=1

\documentclass[11pt]{article}

\usepackage[final]{acl}

\usepackage{times}
\usepackage{latexsym}

\usepackage[T1]{fontenc}

\usepackage[utf8]{inputenc}

\usepackage{microtype}

\usepackage{inconsolata}

\usepackage{graphicx}

\usepackage{xurl}

\usepackage{preamble}

\title{Into the \textit{Unknown Unknowns}: Engaged Human Learning \\
through Participation in Language Model Agent Conversations}

\author{
 \textbf{Yucheng Jiang\textsuperscript{1*}}\quad
 \textbf{Yijia Shao\textsuperscript{1*}}\quad\textbf{Dekun Ma\textsuperscript{2}}\quad\textbf{Sina J. Semnani\textsuperscript{1}}\quad
 \textbf{Monica S. Lam\textsuperscript{1}}
\\
 \textsuperscript{1}Stanford University \quad
 \textsuperscript{2}Yale University\\
\texttt{\{yuchengj, shaoyj\}@cs.stanford.edu} \\ \texttt{dekun.ma@yale.edu}, ~
\texttt{\{sinaj, lam\}@cs.stanford.edu}\\
}

\begin{document}
\maketitle
\def\thefootnote{*}\footnotetext{Equal Contribution}\def\thefootnote{\arabic{footnote}}
\begin{abstract}
While language model (LM)-powered chatbots and generative search engines excel at answering concrete queries, discovering information in the terrain of \textit{unknown unknowns} remains challenging for users. 
To emulate the common educational scenario where children/students learn by listening to and participating in conversations with their parents/teachers, we create \textbf{Collaborative STORM} (\textbf{\system}). \footnote{Our resources and code are released at \url{https://github.com/stanford-oval/storm}.}
Unlike QA systems that require users to ask all the questions, \system lets users observe and occasionally steer the discourse among several LM agents. The agents ask questions on the user's behalf, allowing the user to discover \textit{unknown unknowns} serendipitously. To facilitate user interaction, \system assists users in tracking the discourse by organizing the uncovered information into a dynamic mind map, ultimately generating a comprehensive report as takeaways. For automatic evaluation, we construct the \data dataset by collecting real information-seeking records with user goals. \system outperforms baseline methods on both discourse trace and report quality. In a further human evaluation \footnote{A live research preview is now available publicly at \url{https://storm.genie.stanford.edu}}, 70\% of participants prefer \system over a search engine, and 78\% favor it over a RAG (Retrieval Augmented Generation) chatbot.

\end{abstract}

\section{Introduction}
\begin{figure}[t]
    \centering 
    \resizebox{\columnwidth}{!}{%
    \includegraphics{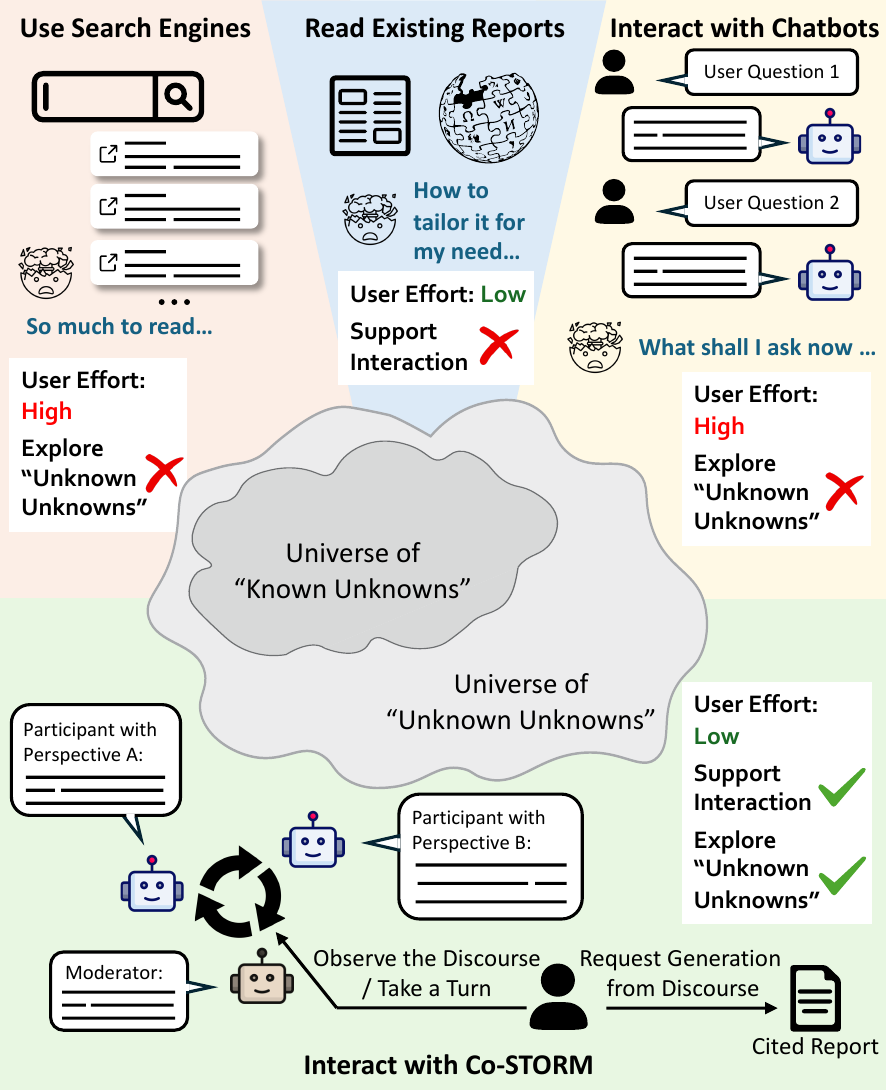}
    }
    \caption{\textbf{Comparison of different paradigms for learning and information seeking.} \system enables humans to observe and participate in a collaborative discourse among LM agents with different roles. Users can request the system to generate a full-length cited report based on the discourse history and the information collected.}
    \label{fig:compare_paradigm}
\vspace{-1.5em}
\end{figure}

Recent advancements in language models (LMs) \citep{bai2022constitutional, gpt4technicalreport, geminiteam2024gemini} and retrieval-augmented generation (RAG)~\citep{lewis2021retrievalaugmented} have led to more capable chatbots and emerging generative search engines~\citep{liu-etal-2023-evaluating}. 
Compared to traditional search engines and information retrieval (IR) models~\citep{Robertson1977THEORIESAM}, these systems fulfill user queries by generating direct responses, effectively addressing \textit{known unknowns}, where users are aware of their information needs. 

However, a gap remains in using these systems for complex information-seeking scenarios, such as academic research, market analysis, and decision-making, where the system should expose users to their \textit{unknown unknowns} to facilitate knowledge discovery. While the concept of \textit{unknown unknowns} originally referred to unexpected risks in the military, it is linked to the serendipitous discovery of information in the information research context~\citep{foster2003serendipity,agarwal2015towards}. Specifically, 
\citet{kirzner1997entrepreneurial} directly contrasts such discovery (``the realization that one had overlooked something in fact readily available'') with successful search (``the deliberate production of information which one knew one had lacked'').


Prior work on automated expository writing attempts to help readers  navigate the terrain of \textit{unknown unknowns} by curating information from various sources into unified articles with substantial breadth and depth~\citep{shen2023summarization}. 
In particular, the STORM writing system demonstrates that LMs paired with search engines can automatically generate a high-quality draft of Wikipedia-like articles on arbitrary topics~\citep{shao2024assisting}. However, producing the static report as the final outcome, STORM does not support any user interaction which is crucial in complex information seeking 
where there is no single, gold query, but queries evolve dynamically towards a goal~\citep{Bates1989TheDO}. This dynamic and exploratory nature makes designing assistance systems challenging. Traditional search engines and RAG chatbots passively react to users' search queries or conversational questions, often inducing echo chamber effects \citep{sharma2024generative} or high cognition load as users with limited prior knowledge may even struggle to formulate questions~\citep{kuhlthau1991inside,belkin1982ask}.

To surface \textit{unknown unknowns} and better support user interaction, we propose \textbf{Collaborative STORM} (\textbf{\system}), an information-seeking assistance system that supports collaborative discourse among users and multiple LM agents. Unlike the one-question-one-answer mode of interaction, \system allows users to learn by observing and participating occasionally in the discourse, emulating a common educational scenario~\citep{nussbaum2008collaborative}. 
To facilitate a thought-provoking discourse and serendipitous discovery, \system simulates two agent types grounded in the search engine: \textit{\experts} who participate by asking or answering questions with different perspectives and a \textit{\moderator}, a non-expert who knows enough to ask good questions and steers the discourse. The user can jump in at any time to steer the discourse and inject questions and opinions according to their interest.  \system maintains a dynamic, hierarchical \textit{mind map} to ensure they can easily follow and engage \citep{buzan1974noting}. 
Upon the conclusion of the discourse, users can request the system to generate a cited report based on the mind map. 



For evaluation, we introduce WildSeek, a dataset of topics and user goals from real users engaged in complex information-seeking across multiple domains. We propose automatic metrics to assess both discourse trace and final report quality. Our results show that \system significantly outperforms RAG chatbots in surfacing in-depth and serendipitous information while providing a more engaging learning experience.

We further conduct a human evaluation by inviting 20 users with diverse backgrounds to compare \system with a search engine and a RAG chatbot. 70\% preferred \system over the search engine, and 78\% preferred it over the RAG chatbot for the overall information-seeking experience. Participants find that \system facilitates serendipitous discovery and requires less mental effort.


Our main contributions include:
\begin{itemize}
    \item {We propose \system, a novel system that combines collaborative discourse emulation, human interaction, and information organization to assist learning and complex information seeking.}
    \item {We construct the \data dataset from real-world human information-seeking records to evaluate information-seeking assistance tools.}
    \item {Results from both automatic and human evaluation show that \system can help humans discover \textit{unknown unknowns} with less mental effort required.}
\end{itemize}

\section{Complex Information Seeking}

\subsection{Problem Formulation}
\label{sec:task_setup}

\citet{pirolli2009powers} defines complex information-seeking as part of the broader sensemaking process, involving collecting, sifting, understanding, and organizing information from large collections to generate a knowledge product. Prevalent in domains such as investigative journalism, scientific research, and market analysis, this task has the following properties: (1) it requires seeking information from \textit{multiple sources} to address various facets of a topic rather than retrieving a document that best matches a query; (2) it involves \textit{ongoing user interaction} rather than processing a single query; (3) it produces \textit{report-like curated information product} rather than a single short-form answer. As shown in \reftab{table:setup_comparison}, 
none of the existing information-seeking assistance systems~\citep[e.g.,][]{Robertson1977THEORIESAM,chen-etal-2017-reading,reddy-etal-2019-coqa,shao2024assisting} can  fully support this task.

\begin{table}
\resizebox{\columnwidth}{!}{%
\centering
\begin{tabular}{lccc} 
\toprule
        & \begin{tabular}[c]{@{}c@{}}\textbf{Multiple}\\\textbf{Sources}\end{tabular} & \begin{tabular}[c]{@{}c@{}}\textbf{Ongoing}\\\textbf{Interact}\end{tabular} & \begin{tabular}[c]{@{}c@{}}\textbf{Curated}\\\textbf{Report}\end{tabular}  \\ 
\midrule
Information Retrieval & $\xmark$                                                                      & $\xmark$                                                                                           & $\xmark$                                                                     \\
Single-Turn QA       & $\cmark$                                                                      & $\xmark$                                                                                           & $\xmark$                                                                     \\
Conversational QA     & $\cmark$                                                                      & $\cmark$                                                                                           & $\xmark$                                                                     \\
Report Generation     & $\cmark$                                                                      & $\xmark$                                                                                           & $\cmark$                                                                     \\ 
\midrule
\textbf{\system}         & $\cmark$                                                                      & $\cmark$                                                                                           & $\cmark$                                                                     \\
\bottomrule
\end{tabular}
}
\caption{Comparison of different information-seeking assistance systems.}
\label{table:setup_comparison}
\vspace{-0.5em}
\end{table}

Given a user with an initial topic of interest $t$ and an initial goal $g$, and a repository of information $R$, the task is to {\em interact} with the user and write a custom long-form report tailored to the users' interest; the report is a sequence of sentences $\mathcal{S}=s_1s_2...s_n$, with each sentence citing a set of retrieved information $\mathcal{I} \subset \mathcal{R}$,  for the sake of verifiability.

\subsection{\data: An In-the-Wild Information Seeking Dataset}
\label{sec:task_data}
\begin{table}
\centering
\resizebox{0.9\columnwidth}{!}{%
    
    \begin{tabularx}{\columnwidth}{X}
    \toprule
        \small{\textbf{Domain}: Economics} \\
        \midrule[0.5pt]
        \small{\textbf{Topic:} Development of a Shared Trading Currency to Facilitate International Trade} \\
        \midrule[0.5pt]
        \small{\textbf{Goal:} Investigate how a new shared currency could eliminate transaction costs and boost GDP among member countries.} \\
    \bottomrule
    \end{tabularx}
}
    \caption{A sample data point in the \data dataset for studying complex information-seeking tasks; the topic and goal are provided by users on the publicly available STORM website, the domain is assigned manually.}
    \label{tab:data_point}
\vspace{-1em}
\end{table}

To study users’ interests in complex information-seeking tasks in the wild,
we utilized data collected from the open-source STORM web application\footnote{Available at \url{https://storm.genie.stanford.edu}}, which generates comprehensive long-form reports based on users’ specified topics of interest and goals for using the site. 
Each data point is a pair comprising a topic and the user's goal. 
To improve the quality of the dataset, we retain only those data points that are well motivated by applying rule-based filtering followed by binary classification using an LM (\texttt{gpt-4o-2024-05-13}). Next, we use the same LM to predict the taxonomy class of each topic, followed by manual review and refinement. Finally, we downsample the data to create a dataset with 100 data points across 24 different domains. \reftab{tab:data_point} shows a sample data point from the dataset; further details about the dataset are in Appendix~\ref{appendix:dataset}.


\section{Method}
\label{sec:method}
\begin{figure*}[t!]
    \centering 
    \resizebox{0.95\textwidth}{!}{%
    \includegraphics{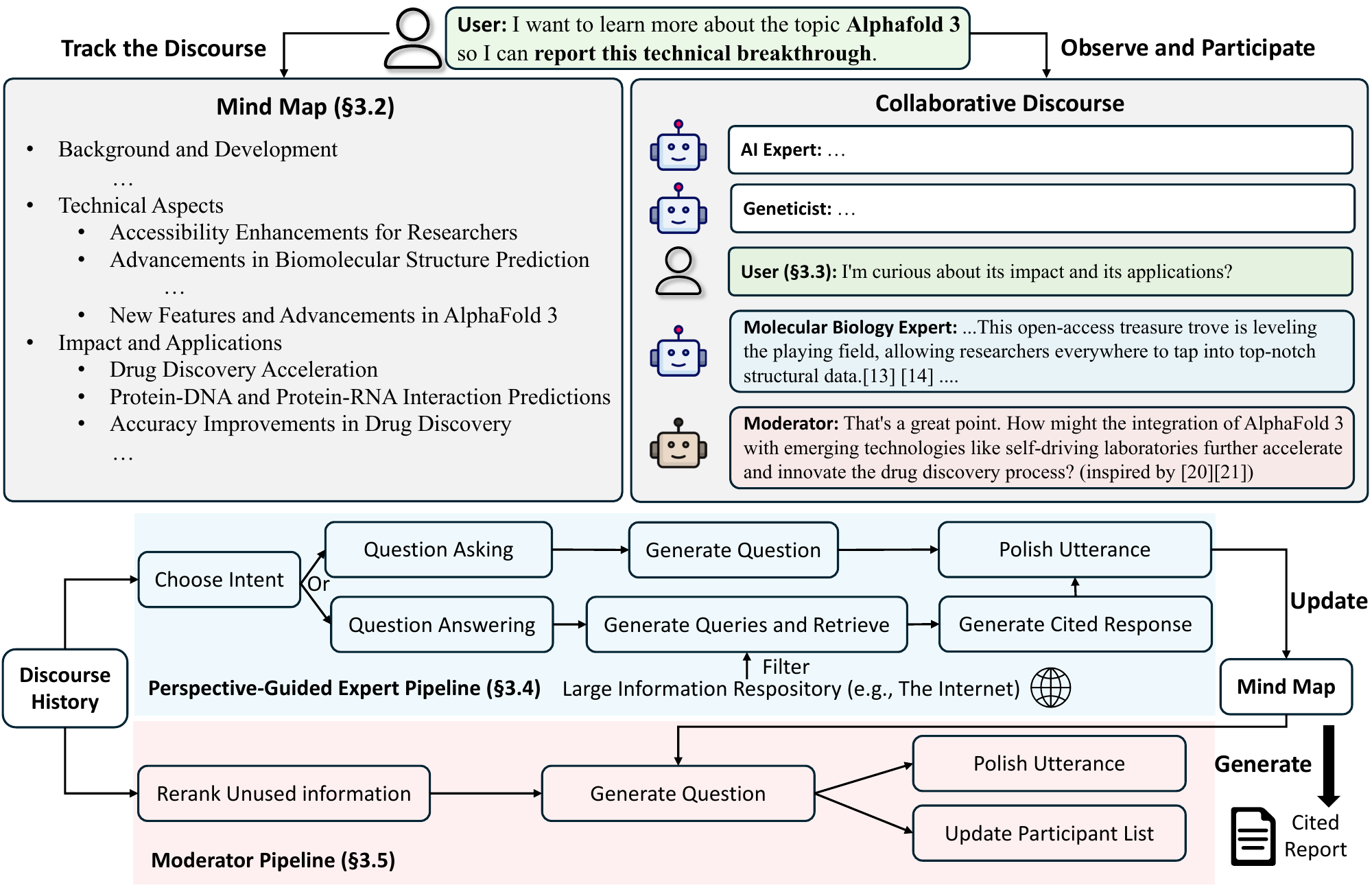}
    }
    \caption{\textbf{Overview of \system.} \system emulates a collaborative discourse among the user, simulated perspective-guided \experts, and a simulated \moderator. It maintains a dynamically updated mind map (\refsec{sec:mind_map}) to help user track and engage in the discourse (\refsec{sec:human_steering_mode}) . The simulated \expert is prompted to determine the utterance intent based on discourse history and generate a question or an answer grounded in the Internet (\refsec{sec:simulate_participant}). The simulated \moderator is prompted with unused information and the mind map to generate a new question to automatically steer the discourse (\refsec{sec:simulate_moderator}). The mind map can be used to generate a full-length cited report as takeaways. Complete discourse transcript and the associated report are detailed in Appendix~\refsec{sec:alphafold_discourse} and~\refsec{sec:alphafold_report}.}
    \label{Fig.method}
\vspace{-1em}
\end{figure*}

\textit{``Tell me and I forget. Teach me and I remember. Involve me and I learn.''}
— Benjamin Franklin

\subsection{Collaborative Discourse Protocol}
\label{sec:discourse_protocol}
\citet{nussbaum2008collaborative} emphasizes the importance of collaborative discourse in fostering deep understanding and critical thinking in human learning. Since it is difficult to assemble a group of human experts for collaborative discourse on any topic at any time, we propose \system (\reffig{Fig.method}) to emulate this process with multiple LM agents to assist human information seeking and learning. Formally, the collaborative discourse, $\mathcal{D}=\{u_1, u_2, ..., u_n\}$, consists of turn-based textual utterances $u_i$ from one of three roles: 
the \textit{user} (\refsec{sec:human_steering_mode}), \textit{\experts} with diverse perspectives (\refsec{sec:simulate_participant}), and a \textit{\moderator} guiding the discourse and injecting questions (\refsec{sec:simulate_moderator}). The discourse begins with $N$ \experts, $\mathcal{P}=\{p_1, ..., p_N\}$, discussing the topic $t$ for one turn per \expert to warm up the conversation. \system dynamically maintains a mind map (\refsec{sec:mind_map}) to track the discourse and construct shared knowledge between the user and the system.

\noindent
\textbf{Utterance Intent}\quad
Inspired by the utterance intent taxonomy for information-seeking conversations proposed by \citet{Qu_2019}, we associate each agent utterance $u_i$ with an agent intent type $a_i$, where $a_i$ can be one of the following: 
\begin{itemize}
\item
\intent{Original Question} initiates a new question 
\item
\intent{Information Request} seeks additional information from the prior utterance
\item
\intent{Potential Answer} offers a possible answer to a previously posed question 
\item
\intent{Further Details} provides supplementary information to a previous answer
\end{itemize}
We group \intent{Original Question} and \intent{Information Request} as \textit{question-asking} and the other two  intents as \textit{question-answering}.


\noindent
\textbf{Initiative Management}\quad
\citet{traum2003issues} underscores the necessity of discourse management in multiparty dialogues. 
While existing systems support either just user initiatives (\eg, QA systems) or just agent initiatives (\eg, STORM), \system adopts a mixed-initiative approach. When the user actively engages, the system continues the discourse based on the user's question or argument, allowing for a more targeted discussion. Otherwise, the system automatically generates the next turn. The user controls who takes the initiative, as \system allows the user to take a turn anytime.

\noindent
\textbf{Turn Management}\quad
If the user does not take the turn at timestamp $i$, \system needs to determine which LM agent should generate the next utterance $u_i$. Its protocol is to let different \experts, $p_1, ..., p_N$, take turns in sequence. To prevent them from only expanding on the same point, upon observing $L$ consecutive turns of expert responses with intents being either \intent{Potential Answer} or \intent{Further Details}, the system asks the \moderator to intervene.
In~\refsec{sec:automatic_evaluation_results}, we analyze the benefit of this protocol design.


\subsection{Tracking the Discourse with a Mind Map}
\label{sec:mind_map}

Having shared knowledge or a shared conceptual space is critical for collaboration~\citep{roschelle1995construction}. To help users track the discourse and reduce their cognition load, \system uses a tree-structured \textit{mind map} $\mathcal{M}$ to dynamically organize collected information in the discourse $\mathcal{D}$. 
Specifically, $\mathcal{M} = (\mathcal{C}, \mathcal{E})$ is a hierarchical organization of concepts $\mathcal{C}$, where its directed edges $\mathcal{E}$ characterize latent parent-child relationships among topics (\eg, in \reffig{Fig.method}, ``Drug Discovery Acceleration'' is a subtopic of ``Impact and Applications''). Each concept $c\in\mathcal{C}$ is associated with a subset of retrieved information $I^{c}\subset\mathcal{I}$. To ensure $\mathcal{M}$ is an intent-driven organization of information, each piece of information is also associated with the question that leads to its retrieval.

\system dynamically updates the mind map through two operations, \texttt{insert} and \texttt{reorganize}. \texttt{Insert} places a piece of information under the most appropriate concept by first deriving a set of candidate concepts using semantic similarity between its associated question and each concept in $\mathcal{C}$, then prompting the LM to choose the final placement. When a concept $c$ has more than $K$ pieces of information, $\mathcal{M}$ triggers the \texttt{reorganize} operation. \texttt{Reorganize} prompts the LM to generate a list of new subtopic names under $c$, and applies \texttt{insert} to place each piece of information associated with $c$ in the subtree rooted at $c$. After expansion, \system adopts a bottom-up cleaning process to iteratively delete concepts with no supporting information and collapse concepts with only one subtopic. More details on \texttt{insert} operation are included in Appendix~\ref{sec:mind_map_insertion}.

\subsection{User Participation}
\label{sec:human_steering_mode}
When the user injects an utterance $u$, \system uses $u$ as the query to retrieve information to prompt the LM to obtain an updated list of \experts, $\mathcal{P}'$. 
Following this update, the system switches back to the auto-steering mode where the \expert or the \moderator takes turns according to the turn management protocol introduced in~\refsec{sec:discourse_protocol}. Once the user is satisfied with the discourse, \system generates the final report $\mathcal{S}$ as the curated information product of the collaborative discourse. This report is generated using the mind map $\mathcal{M}$ as the outline and the retrieved information $I^{c}$ associated with each concept $c$ to generate the report section by section.

\subsection{Simulating the Roundtable Participant}
\label{sec:simulate_participant}
Following STORM~\citep{shao2024assisting}, which uses perspective-guided question asking to improve the question diversity and quality, \system personalizes simulated \experts with different expertise to represent different perspectives. \system retrieves the background of topic $t$ with a search query and gives it to an LM to generate the \expert list $\mathcal{P}=\{p_1, ..., p_N\}$. For example, for the topic ``AlphaFold3'' in \reffig{Fig.method}, the LM suggests an ``AI Expert'', a ``Geneticist'', and a ``Molecular Biology Expert'' to participate in the discourse.

If there is no interruption by the user or the \moderator, each \expert $p_j$ sequentially takes turns with the following procedure: 
To generate an utterance from an expert $p$ in turn $i$, 
(1) The LM is prompted to choose the intent $a_i$ based on the discourse history $\{u_1, ..., u_{i-1}\}$ and the \expert's perspective $p$. (2) If  the intent $a_i$ is \intent{Potential Answer} or \intent{Further Details}, we prompt the LM to generate a search query $q$, retrieve information with a search engine~\footnote{Following \citet{shao2024assisting}, when retrieving information with a search engine, we apply rule-based filtering according to the Wikipedia guideline \url{https://en.wikipedia.org/wiki/Wikipedia:Reliable_sources}.}, and  generate a response with citations; otherwise, we prompt the LM to directly generate a question based on the discourse history.  
(3) We use the LM to polish the utterance to make it more chatty and engaging.

\subsection{Simulating the Moderator}
\label{sec:simulate_moderator}

If all the simulated participants are experts, we discover that the discourse tends to consist mostly of utterances with  intent \intent{Further Details}, leading to repetition and niche discussions. The moderator plays an important role of injecting new directions into the discourse.  
To generate the \moderator's utterance, \system instructs the LM to generate informed questions based on the uncited sources retrieved since the last \moderator turn. To choose among the many uncited sources, the \moderator reranks each piece of information $i$ based on the similarity to the topic $t$ and the dissimilarity to its associated question $q$. 
Formally, the reranking score is 
\begin{equation}\nonumber
    \operatorname{cos}(\mathbf{i}, \mathbf{t})^{\alpha} (1- \operatorname{cos}(\mathbf{i}, \mathbf{q}))^{1-\alpha},
\vspace{-0.5em}
\label{eq:rerank_score}
\end{equation}
where $\mathbf{i}, \mathbf{t}, \mathbf{q}$ are corresponding text embeddings and $\alpha$ is a hyperparameter. This reranking function prioritizes information that does not directly answer the original question but is relevant to the topic $t$. \system concatenates these reranked sources along with concept names in $\mathcal{C}$ to avoid repetitive concepts. This combined context is used to prompt the LM to generate the question for the \moderator turn and an updated list of \experts, $\mathcal{P}'$.



\section{\system Implementation}

\begin{table*}
\centering
\resizebox{\textwidth}{!}{%
\begin{tabular}{lccccc!{\vrule width \lightrulewidth}ccc} 
\toprule
                        & \multicolumn{5}{c!{\vrule width \lightrulewidth}}{\textbf{Report Quality}} & \multicolumn{3}{c}{\textbf{Question-Answering Turn Quality}}  \\
                        & Relevance & Breadth & Depth & Novelty & Info Diversity                                    & Consistency & Engagement & \# Unique URLs      \\ 
\midrule
RAG Chatbot             & 3.57     & 3.50    & 3.26  & 2.44        & 0.595                                  & 4.37        & 4.13       & 2.94                           \\
STORM + QA              & 3.61     & 3.61    & 3.43  & 2.50                    &0.592                      & 4.34        & 4.11       & 2.89                         \\ 
\midrule
\textbf{\system}                & \textbf{3.78}     & \textbf{3.79}    & \textbf{3.77}\dag   & \textbf{3.05}\dag      & \textbf{0.602}                                    & \textbf{4.40}\dag        & \textbf{4.33}\dag       & \textbf{6.04}\dag                     \\
~ w/o Multi-Expert & 3.73     & 3.75    & \textbf{3.77}  & 2.93                           & 0.589               & \textbf{4.40}        & 4.32       & 5.91                    \\
~ w/o Moderator         & 3.56     & 3.69    & 3.41  & 2.89                       & 0.577                   & 4.39        & 4.28       & 5.67                     \\
\bottomrule
\end{tabular}
}
\caption{Automatic evaluation results for report quality and the quality of question-answering turns in the discourse with simulated users. Ablations are included as follows: “w/o Multi-Expert” denotes 1 expert and 1 moderator, and “w/o Moderator” denotes $N$ experts and 0 moderator.\dag~denotes significant differences ($p<0.05$) from a paired $t$-test between \system and both baselines. The rubric grading uses a 1-5 scale. All scores reported are the mean values.} 
\vspace{-1em}
\label{table:report_grading}
\end{table*}

The LM component of \system is implemented using zero-shot prompting via the DSPy framework \citep{khattab2023dspy} and \texttt{gpt-4o-2024-05-13} (see full prompts in Appendix~\ref{appendix:rubric}). We ground \system on the Internet using the You.com search API\footnote{\url{https://documentation.you.com/api-reference/search}} although the system is compatible with other search engines or IR systems. Hyperparameters $N$, $K$, $L$, $\alpha$ are set to 3, 10, 2, and 0.5, respectively. The text embeddings in \refequ{eq:rerank_score} are obtained from \texttt{text-embedding-3-small}. We set the LM temperature as 1.0 and top\_p as 0.9 for all experiments. For human evaluation, we develop a web application (\reffig{Fig.human_eval_demo}) for users to interact with \system in real time.

\section{Automatic Evaluation}
\label{sec:auto_eval}

Automatic evaluation enables scalable testing and allows for consistent simulation of user behavior. We compare \system with the following baselines: (1) \textbf{RAG Chatbot}, a baseline that retrieves information from the search engine and interacts with the user through a one-question-one-answer paradigm. (2) \textbf{STORM + QA}, a baseline that uses the STORM framework~\citep{shao2024assisting} to generate a report for a given topic to provide general information. It then allows the user to ask follow-up questions and provides corresponding answers retrieved with the search engine.

\subsection{Evaluation Setup}
We use the \data dataset where each data point consists of an initial topic $t$ and goal $g$. 
We simulate the user 
with an LM (\texttt{gpt-4o-2024-05-13}) prompted with $t$, $g$, the discourse history $\mathcal{D}$, and the instruction for question generation. To ensure a fair comparison, we terminate the information-seeking session once it reaches 30 search queries for \system and both baselines. For all methods, the final report is generated using the two-stage approach of outline generation followed by section-by-section article generation, as proposed by STORM~\citep{shao2024assisting}, based on the interaction history. We evaluate the system quality by assessing the final report and the interaction history (\ie, discourse) with the automatic metrics defined in~\refsec{sec:eval_metrics}.

\subsection{Automatic Metrics}
\label{sec:eval_metrics}
\noindent
\textbf{Report Quality}\quad
We evaluate the final report on four aspects, \textit{Relevance}, \textit{Broad Coverage (Breadth)}, \textit{Depth}, and \textit{Novelty}, as indicators of the quality of the whole information-seeking process.\footnote{The same four aspects are used in human evaluation.} We employ Prometheus 2 \citep{kim2024prometheus}, a 7B evaluator LM, to score the report based on a 5-point rubric. To further quantify the diversity of the collected information, we also report the \textit{Information Diversity} as the average pairwise dissimilarity of $\mathcal{I}$,
\begin{equation}\nonumber
    1-\frac{\sum_{i,j\in\mathcal{I},i\neq j}\operatorname{cos}(\mathbf{i}, \mathbf{j})}{|\mathcal{I}|(|\mathcal{I}|-1)},
\end{equation}
where $\mathbf{i},\mathbf{j}$ are corresponding text embeddings obtained from OpenAI's \texttt{text-embedding-3-small}.


\noindent
\textbf{Discourse Quality}\quad
Since the discourse itself is valuable for human learning, we also evaluate the discourse trace using a 5-point rubric to grade each turn. This grading assesses \textit{Novelty}, \textit{Intent Alignment}, and \textit{No Repetition} for question-asking utterances (\ie, utterances with the intent \intent{Original Question} or \intent{Request Information}). For question-answering utterances that provide information
, we assess \textit{Consistency} and \textit{Engagement}. We also report the number of unique cited URLs in these utterances to indicate information diversity at the turn level. 
Both the rubrics for report evaluation and utterance evaluation are included in Appendix~\ref{appendix:rubric}.

\label{sec:results}

\subsection{Automatic Evaluation Results}
\label{sec:automatic_evaluation_results}
\reftab{table:report_grading} presents the evaluation results for report quality and the quality of question-answering turns in the discourse. The question-answering turns and the final report are the primary sources for human learning when they interact with \system. STORM + QA considers multiple perspectives in researching the given topic, indeed leading to improved performance across all four grading dimensions of the report quality compared to the RAG Chatbot. However, \system outperforms it, particularly in the \textit{Depth} and \textit{Novelty} aspects, by simulating collaborative discourse with multiple agentic roles, akin to a thought-provoking round table discussion. 
For discourse quality, the question-answering turns in \system significantly outperform both baselines in terms of \textit{Consistency} and \textit{Engagement}. This improvement is attributed to collaborative discourse setup, where the LM is prompted to generate the answer only when the retrieved information matches the current question according to the discourse history (see Listing~\ref{lst:participants_prompt}). The utterance polishing step (see \reffig{Fig.method}) also helps as it serves as a self-improving mechanism.

\begin{figure}[t]
    \centering 
    \resizebox{\columnwidth}{!}{%
    \includegraphics{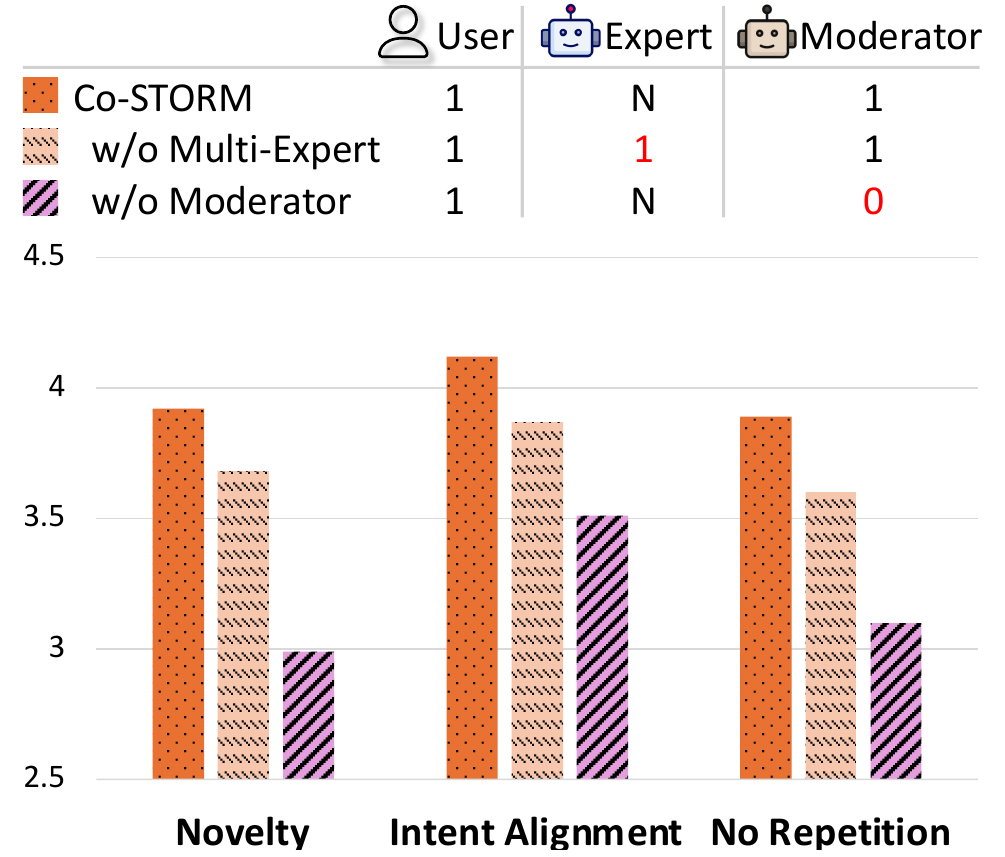}
    }
    \caption{Rubric grading results for question-asking turn quality in automatic evaluation with simulated users.}
    \label{fig:question_quality}
\vspace{-1em}
\end{figure}


\subsection{Ablation Studies}
As discussed in~\refsec{sec:method}, a major innovation of \system is the orchestration of two types of LM agents. 
To assess the benefit, we compare \system with two ablations: (1) without multiple \experts with different perspectives (``w/o Multi-Expert''), \ie, only a single \expert and a \moderator, and (2) multiple experts but no \moderator steering the discourse (``w/o Moderator''). As shown in~\reftab{table:report_grading}, the ablated systems perform worse than the full system across all metrics in both report and question-answering turn quality. Notably, removing the \moderator has a greater negative impact than reducing the number of \experts.

A key feature of \system is that LM agents can ask questions on the user's behalf. As shown in~\reffig{fig:question_quality}, the advantage of \system's multi-agent design becomes clearer when inspecting the question-asking turns. \textit{Having just one \expert and one \moderator can already provide most of the benefits.} Importantly, the \moderator role in \system raises questions based on unused information about the topic---such a role represents somebody with a much larger \textit{known unknowns}, effectively steering the discourse to help users discover more in the space of their \textit{unknown unknowns}.

Another key innovation of \system is the dynamic mind map. 
We include controlled experiments on mind map quality in Appendix~\ref{sec:mind_map_insertion}.


\section{Human Evaluation}
\label{sec:human_eval}
Human evaluation is essential for assessing systems designed for collaborative discourse, as it captures the complexities of human interaction, reflects familiar real-world interactions, and provides critical insights into the system’s effectiveness. We compare \system with two baselines: (1) \textbf{RAG Chatbot} as detailed in section \refsec{sec:auto_eval}. (2) Traditional \textbf{Search Engine}.
\subsection{Evaluation Setup}
We conduct an IRB-approved human evaluation to compare \system with RAG Chatbot and Search Engine by recruiting 20 volunteers on the Internet. Participants are randomly split into two groups: one compared \system with Google Search, while the other group compared it with the RAG Chatbot. Participants are asked to seek information on two topics, one on each system, from the same domain and sharing the same goal. Note that we mitigate topic familiarity bias by using two different topics within the same domain.
We have prepared five different domains, with each assigned to 2 users in each group. To counterbalance, one user starts with \system and switches to the baseline for the other topic, and vice versa. 


After seeking information for each topic, participants are instructed to rate their experience based on four grading aspects defined in \refsec{sec:eval_metrics} (\textit{Relevance}, \textit{Breadth}, \textit{Depth}, \textit{Novelty/Serendipity}), using a 5-point Likert scale. After completing both tasks, participants are asked to provide pairwise preferences regarding the \textit{required effort}, \textit{user engagement}, \textit{addressing echo chamber issues}, and \textit{overall experience}. We also collect open-ended feedback and allow participants to optionally leave comments on each discourse turn and the mind map snapshots when interacting with \system. More details on the human evaluation are included in Appendix~\ref{appendix:human_eval}.
\begin{table*}[!h]
\centering
\resizebox{\textwidth}{!}{%
\begin{tabular}{lcccc!{\vrule width \lightrulewidth}cccc} 
\toprule
            & \multicolumn{4}{c!{\vrule width \lightrulewidth}}{\textbf{\system v.s. Search Engine}} & \multicolumn{4}{c}{\textbf{\system v.s. RAG Chatbot}}  \\
            & Search Engine & \system & Win \% (Lose \%) & $p$-value                                           & RAG Chatbot & \system & Win \% (Lose \%) & $p$-value             \\ 
\midrule
Relevance   & 3.90          & \textbf{4.00}          &   30\% (30\%)       & 0.758                                             & 3.89        & \textbf{4.22}          &     33\% (0\%)     & 0.081               \\
Breadth     & 3.60          & \textbf{4.10}          &   50\% (10\%)       & 0.096                                             & 3.11        & \textbf{4.22}          &     67\% (0\%)     & 0.013               \\
Depth       & 3.10          & \textbf{4.00}          &    60\% (10\%)      & 0.081                                             & 3.11        & \textbf{4.00}          &    56\% (33\%)      & 0.069               \\
Serendipity & 2.70          & \textbf{3.90}          &   70\% (10\%)       & 0.030                                             & 2.78        & \textbf{3.78}          &     67\% (0\%)     & 0.009               \\
\bottomrule
\end{tabular}
}
\caption{Human ratings on different aspects of the information-seeking experience with \system and Search Engine (n=10) and with \system and RAG Chatbot (n=9)\protect\footnotemark. The ratings are given on a scale from 1 to 5 with 3 as ``Average''. We report the win rate of \system in pairwise comparison and the $p$-value in a paired $t$-test.}
\label{table:human_eval}
\vspace{-1em}
\end{table*}

\footnotetext{One participant in the \system v.s. RAG Chatbot group submitted the rating but did not leave a usage record, so we excluded this data point from the aggregated results.}

\begin{figure}[t]
    \centering 
    \resizebox{\columnwidth}{!}{%
    \includegraphics{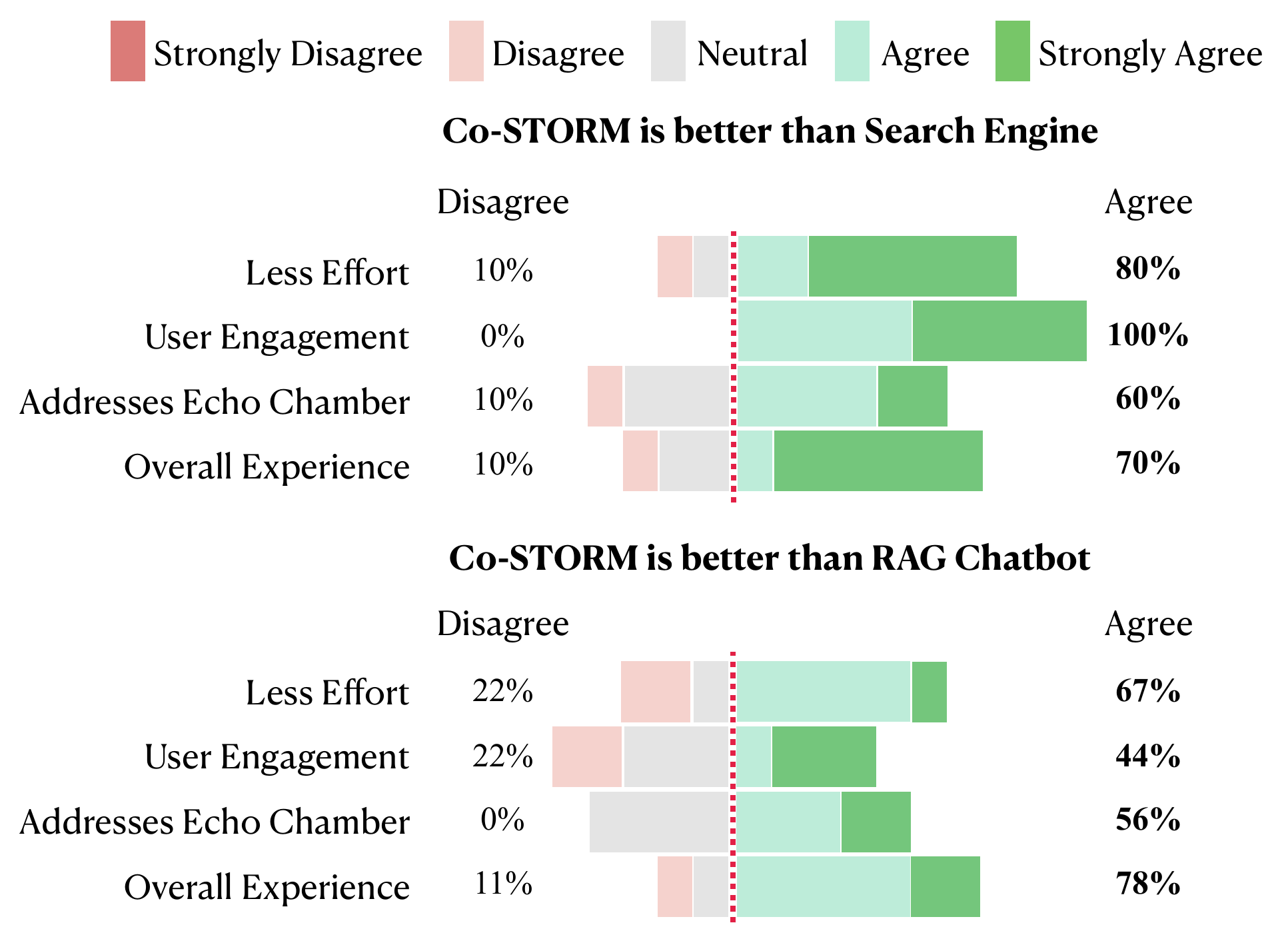}
    }
    \caption{Survey results of the pairwise comparison (\ie, agreement on whether \system is better than Search Engine/RAG Chatbot) in human evaluation.}
    \label{fig:human_eval_pref}
\vspace{-1em}
\end{figure}

\subsection{Human Evaluation Results}
\reftab{table:human_eval} shows the human rating results and \reffig{fig:human_eval_pref} shows the pairwise comparison results.
\textbf{\system helps users find broader and deeper information relevant to their goals.}\quad
Participants found that \system uncovers information with greater breadth and depth compared to the search engine and the RAG Chatbot. Specifically, \system is rated strictly higher in \textit{Breadth} by 50\% of the participants and strictly higher in \textit{Depth} by 60\% of the participants than the search engine. Compared to the RAG Chatbot, \system receives strictly higher scores in \textit{Breadth} from 67\% of participants and in \textit{Depth} from 56\% of participants. This finding aligns with the automatic evaluation results shown in~\reftab{table:report_grading}. While helping users discover more information, \system remains aligned with their goals, as participants also rated \system higher in \textit{Relevance} compared.


\noindent
\textbf{\system provides more serendipitous information with less mental effort required.}\quad
Participants found that \system requires less effort, better mitigates the echo chamber issue, and provides a better overall experience. In a more fine-grained evaluation, participants evaluated 32\% of \system’s total utterances, rating 89\%
of them as effectively ``steering the discourse towards a new and interesting direction''. 
One participant noted, ``\system allows for almost full automation and much better understanding as it brings up topics that the user may not even think of''. Moreover, participants found the mind map helpful. 
In total, they evaluated 80 snapshots of the dynamic mind map, 
finding it accurately tracked the discourse 71\%
of the time. One participant remarked, ``\system is so much less mentally taxing for me to use''.


\noindent
\textbf{\system should support more customization}.\quad
Among the 19 participants, 4 noted that the RAG Chatbot better follows instructions that have a clear target and mentioned they expect \system to generate more concise utterances and provide less information in such cases. 
We view dynamically adapting \system to users’ evolving mental states and personalizing their preferences as a meaningful direction for future work.



\section{Related Works}
\noindent
\textbf{Information-Seeking Support in NLP}\quad
NLP research supporting human information-seeking has mainly focused on building question-answering (QA) systems~\citep{chen-etal-2017-reading, lee-etal-2019-latent,dasigi-etal-2021-dataset,levy-etal-2021-open,yuan-etal-2020-interactive-machine}. These works often assume that the answer can be found within a single document~\citep{clark-etal-2020-tydi} or that users can formulate complex queries~\citep{yang-etal-2018-hotpotqa,chen2021open,ahmadvand2023making}, assumptions that do not hold true in complex information seeking~\citep{butler2000learners,booth2009craft,bystrom1995task}.


Some more recent works have proposed long-form QA systems~\citep{xu-etal-2023-critical,xu2024kiwi} and automatic expository writing systems~\citep{balepur-etal-2023-expository,shen2023summarization,shao2024assisting} to synthesize information from multiple sources. 
Some other studies have explored conversational search~\citep{kumar-callan-2020-making,nakamura-etal-2022-hybridialogue}. 
However, these works typically ignore human interaction or only passively answer user questions. We construct a multi-agent system with a human-in-the-loop protocol to support effective user interaction for complex and evolving information needs.

\noindent
\textbf{Multi-Agent Systems}\quad
As LMs advance, a growing body of research explores their use in multi-agent applications~\citep{wu2023autogen,BabyAGI,liu2023dynamic,wang2024unleashing}. Several studies show that multi-agent debate enhances factuality and reasoning compared to using a single LM~\citep{du2023improving,liang2023encouraging}, and cooperative role-playing frameworks improve performance on coding or mathematical benchmarks~\citep{NEURIPS2023_a3621ee9,hong2023metagpt}. While these studies primarily focus on automating tasks, the potential applications extend further. For instance, Generative Agents~\citep{Park2023GenerativeAgents} instantiate an interactive environment with twenty-five LM agents to study emergent social behaviors, and \citet{michael2023debate} show that multi-agent debates help humans supervise model outputs. Our work aligns with these broader applications by constructing a multi-agent system to facilitate human learning.

\noindent
\textbf{Collaborative Discourse for Human Learning}\quad
Collaborative discourse has long been valued in classroom settings for its ability to deepen learners' understanding of concepts, enhance peer learning, and increase engagement 
\citep{nussbaum2008collaborative,osborne2010arguing,kolodner2007roles,chinn2000structure}. Specifically, \citet{nussbaum2008collaborative} argues not all types of collaborative discourse are equally beneficial to students' learning, emphasizing the importance of critical discussion where participants assume different points of view. Furthermore, the facilitator role is important in collaborative discourse, with asking questions and providing complementary information as popular strategies~\citep{onrubia2022assisting}.

\section{Conclusion}
We propose \system, an information-seeking assistance system that emulates collaborative discourse among users and multiple LM agents. By creating an interactive environment where users can both observe and participate, \system enhances learning and the complex information-seeking process. To facilitate automatic evaluation, we construct the \data dataset, which captures the information-seeking needs and intents of real Internet users. Experimental results, including extensive human assessments, show that \system outperforms traditional search engines and RAG chatbots in surfacing \textit{unknown unknowns} for human learning and reducing users' mental effort.

\section*{Limitations}
We design \system to create an immersive human learning experience by enabling humans to participate in LM agent conversations. Despite the advantages demonstrated through both automatic and human evaluations, several limitations remain. First, the system could better tailor the collaborative discourse to the user’s prior knowledge, skipping basic facts for knowledgeable users and introducing concepts progressively for novices. Second, while \system employs an effective discourse management mechanism, users sometimes desire more control over the discourse, including managing \expert perspectives and customizing the utterance length. Third, extending \system to support multiple languages would significantly enhance its usefulness and impact. Although current LMs often possess multilingual capabilities, implementing a multilingual \system requires integrating search engines or retrieval models capable of accessing diverse language sources. Furthermore, managing content across different languages demands robust content moderation and the ability to identify conflicting information to ensure a reliable human learning experience. Finally, compared to the RAG Chatbot, \system has higher latency due to the need to decide the utterance intent and update the mind map. Although the current latency is acceptable for real-time interaction, as demonstrated in human evaluations, further improving the efficiency of the LM system would provide a smoother user experience.



\section*{Acknowledgements}
We thank Omar Khattab, Eric Zelikman, Rose E. Wang, Yen-Jen Wang, and Qingyang Tang for their helpful feedback, and the ACL ARR reviewers for their valuable comments. We also appreciate the 20 participants in our user study who made this work possible. We are grateful to You.com for providing a discounted search API that supported our experiments. Yijia Shao is supported by a Stanford School of Engineering Fellowship. This work is supported in part by the Verdant Foundation and Microsoft Azure AI credits. 

\section*{Ethics Statement}

We build and evaluate our work to strictly adhere to ethical standards. The construction of the WildSeek dataset involves collecting data with users’ explicit approval, and we carefully remove all personally identifiable information. In contrast to creative generation tasks, our tasks generate content that may impact how people perceive information and shape their opinions. We design our system to ground generated content on openly accessible external sources available on the general internet, with proper citations. Our experiments and evaluations ensure the accurate delivery of information and significantly reduce hallucinations. We avoid publishing or posting any generated content without careful examination of information accuracy. We believe there are no data privacy issues as we ground our generated content from information accessible to the general public.

The primary risk of our work is the common bias issues originating from biases present on the general internet. Following \citet{shao2024assisting}, we mitigate this problem by applying rule-based filtering according to the Wikipedia guideline\footnote{\url{https://en.wikipedia.org/wiki/Wikipedia:Reliable_sources}} and incorporating multiple sources. Additionally, in our web application, we have implemented input topic moderation to reject topics that are sensitive, illegal, or potentially violate personal privacy. However, further information processing modules that serve as filters for internet sources and more robust modules to verify the accuracy of information can be implemented. Additionally, our current work only considers generating and retrieving information from English sources. Extending our system to be compatible with multilingual sources and generation will be beneficial.

\bibliography{anthology,custom}

\clearpage
\appendix
\section{Dataset Details}
\label{appendix:dataset}
We constructed \data using a web application\footnote{Our institution's IRB approved the web application} we built that hosts the open-sourced STORM project \citep{shao2024assisting}, as detailed in \refsec{sec:task_data}. User privacy was strictly maintained by explicitly obtaining consent each time users logged into our web application. No personally identifiable information was collected, and the entire dataset was manually reviewed to ensure compliance with this standard. We rejected topics that were illegal, harmful, violent, racist, sexual, non-English, based on personal experience, or contained personal information.

\begin{figure*}[t!]
    \centering 
    \resizebox{0.95\textwidth}{!}{%
    \includegraphics{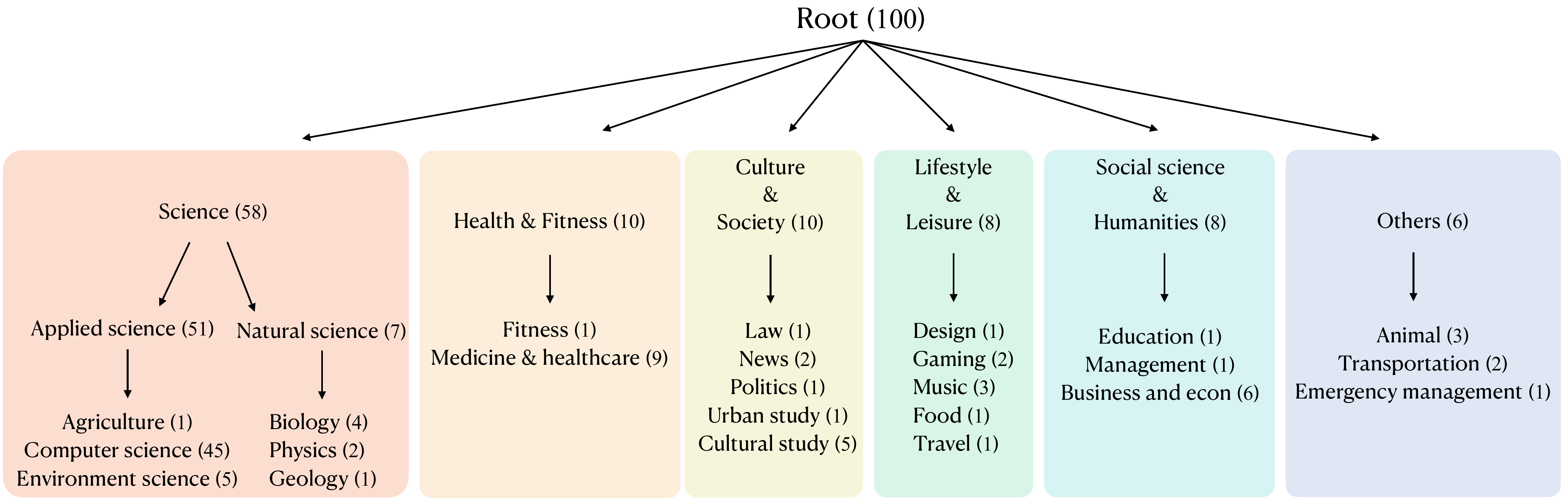}
    }
    \caption{\data taxonomy. The number in the parenthesis denotes the number of data points classified under the corresponding category or its descendants.}
    \label{fig:taxonomy}
\end{figure*}

At the time of dataset construction, 8,777 users accessed our web application, resulting in the collection of 6,608 unique topic and purpose pairs. Participants were sourced from the general internet and all held valid Google accounts, in accordance with our IRB-approved policy. To ensure broad coverage, we conducted topic classification using \texttt{gpt-4o-2024-05-13} and human inspection, and then downsampled the collected data to 100 cases, covering 24 fine-grained categories in 6 domains: Science, Health and Fitness, Culture and Society, Lifestyle and Leisure, Social Science and Humanities, and Others. We applied rule-based filtering to exclude non-informative or trivial information-seeking purposes, and the 100 selected topic-purpose pairs were manually labeled by the authors. \reftab{table:example_tasks} includes example data points from each domain and \reffig{fig:taxonomy} shows the full taxonomy of the \data dataset.

\begin{table*}[!h]
\centering
\resizebox{\textwidth}{!}{%
\begin{tabular}{ll} 
\toprule
\textbf{Domain} & \textbf{Example Task} \\ 
\midrule
Computer Science & 
\begin{tabular}[t]{@{}l@{}} 
\textbf{Topic:} Blockchain anomaly detection using large models \\ 
\textbf{Intent:} To evaluate the effectiveness of large models in detecting anomalies \\ in blockchain systems compared to existing models.
\end{tabular} \\ 
\midrule
Healthcare & 
\begin{tabular}[t]{@{}l@{}} 
\textbf{Topic:} The effects of NMN supplements on human anti-aging \\ 
\textbf{Intent:} To investigate the efficacy and mechanisms of NMN supplements in slowing down \\ or reversing the aging process in humans.
\end{tabular} \\ 
\midrule
Environmental Science & 
\begin{tabular}[t]{@{}l@{}} 
\textbf{Topic:} Utilization of Weather Forecasting for Wind and Solar Energy Assessment \\ 
\textbf{Intent:} To explore advanced methodologies in integrating weather forecast \\ data for optimizing wind and solar energy evaluations.
\end{tabular} \\ 
\midrule
Law & 
\begin{tabular}[t]{@{}l@{}} 
\textbf{Topic:} Recent legal cases in the US involving hardware technology innovations \\ 
\textbf{Intent:} To investigate the legal precedents and implications of hardware technology \\ innovations in the US.
\end{tabular} \\ 
\midrule
Economics & 
\begin{tabular}[t]{@{}l@{}} 
\textbf{Topic:} Development of a Shared Trading Currency to Facilitate International Trade \\ 
\textbf{Intent:} Investigate how a new shared currency could eliminate transaction costs \\ and boost GDP among member countries.
\end{tabular} \\ 
\bottomrule
\end{tabular}
}
\caption{Examples of complex information seeking tasks from the \data dataset.}
\label{table:example_tasks}
\vspace{-0.3em}
\end{table*}

\section{Mind Map \texttt{Insert} Operation}
\label{sec:mind_map_insertion}

\begin{table*}
\centering
\begin{tabular}{lc!{\vrule width \lightrulewidth}cc!{\vrule width \lightrulewidth}cc} 
\toprule
 & \textbf{First-Level} & \multicolumn{2}{c!{\vrule width \lightrulewidth}}{\textbf{Second-Level}} & \multicolumn{2}{c}{\textbf{Third-Level}}  \\
 & \textit{Acc.}        & \textit{Acc.} & Partial \textit{Acc.}                                    & \textit{Acc.} & Partial \textit{Acc.}     \\ 
\midrule
Embedding only & 24.24 & 35.94 & 65.62 & \textbf{35.71} & 57.14 \\
Language Model only & 3.03 & 7.81 & 62.50 & 7.14 & \textbf{71.43} \\
\textbf{\system \texttt{insert}} & \textbf{39.39} & \textbf{51.56} &\textbf{68.75} & \textbf{35.71} & \textbf{71.43} \\
\bottomrule
\end{tabular}
\caption{Controlled experiment results of different mind map insertion methods (\%). A placement is deemed as partially correct if the information is inserted into one of the ancestors of the ground truth placement.}
\label{table:insertion_eval}
\end{table*}

As revealed in human evaluation results (see~\refsec{sec:human_eval}), the mind map is crucial for helping users track the discourse and the collected information. \system dynamically updates the mind map through \texttt{insert} and \texttt{reorganize} operations. In this section, we conduct controlled experiments on different implementations of \texttt{insert} and verify the quality of the mind map updates.

Dynamically organizing collected information into a mind map is challenging. Unlike classic document classification tasks \citep{zhang2024teleclass, zhang2023effective} and recursive summarization tasks \citep{sarthi2024raptor, gao-etal-2023-enabling}, where either the hierarchical organization or the information to be organized is fixed, mind map insertion involves an evolving hierarchical organization of concepts and an incremental set of information. We compare the \system \texttt{insert} (\refsec{sec:mind_map}) with two alternative approaches: (1) The \textbf{Embedding Only} baseline selects the placement with the highest semantic similarity using embedding cosine similarity. (2) The \textbf{Language Model Only} baseline directly prompts an LM to choose the best placement within the given hierarchical organization.



We construct an evaluation dataset for the controlled experiments by leveraging the FreshWiki dataset~\citep{shao2024assisting}, which is a collection of recent, high-quality Wikipedia articles. We use the Wikipedia article outline as the concept hierarchy and require each candidate method to find the best placement for a given citation used in the article. The original placement of the citation in the article is deemed as the ground truth. We apply rule-based filtering to retain articles with up to three levels of hierarchy and English citation sources only. Inserting one cited source back into the outline is considered as one task. After downsampling, we derive a dataset consisting of 111 tasks: 33 from first-level sections, 64 from second-level sections, and 14 tasks from third-level sections.

We report the insertion accuracy in~\reftab{table:insertion_eval}. For tasks where the ground truth placement is in the second or third level, we also consider a placement is partially correct if the information is inserted into one of the ancestors of the ground truth placement and report the partial accuracy. The experimental results show that solely relying on the LM performs poorly as the hierarchical organization can be wide and deep and the performance heavily depends on the quality of concept names. \system \texttt{insert} operation consistently outperforms both baseline approaches.


\section{Full Prompts in \system}
\label{appendix:prompts}
In \refsec{sec:discourse_protocol}, we introduce \system’s collaborative discourse protocol which includes three key roles: the \textit{user}, \textit{experts}, and a \textit{\moderator}. We implement the perspective-guided expert and \moderator pipeline using zero-shot prompting of \texttt{gpt-4o-2024-05-13}. Listing~\ref{lst:participants_prompt} and Listing~\ref{lst:moderator_prompt} documents the full prompts for simulating the \expert and the \moderator respectively. \system uses a hierarchical mind map to track the discourse (\refsec{sec:mind_map}) and the mind map \texttt{insert} operation is detailed in Appendix~\ref{sec:mind_map_insertion}. 
Prompts used for the mind map operations can be found in Listing~\ref{lst:mind_map_prompt}.

\section{Automatic Evaluation Details}
\label{appendix:rubric}
Following \citet{shao2024assisting}, we use the Prometheus model~\citep{kim2024prometheus}, an open-source rubric grading model for evaluating long-form text based on user-defined criteria. For our experiments, we use \texttt{prometheus-7b-v2.0} \footnote{\url{https://huggingface.co/prometheus-eval/prometheus-7b-v2.0}} with its default temperature as 1.0 and top\_p as 0.9, the state-of-the-art version at the time of our experiments. As the model has a limited context window, for report evaluation, we omit references and trim the input text to under 2000 words to fit into the model’s context window, following the practice in \citet{shao2024assisting}; for discourse quality evaluation, we reduce the discourse history length by taking the last 2000 words as context. The report quality evaluation and discourse quality evaluation rubrics can be found in ~\reftab{table:final_report_rubric},~\reftab{table:question_answering_rubric}, and~\reftab{table:question_asking_rubric}.

To assess the quality of the automatic evaluation results in \refsec{sec:automatic_evaluation_results}, we randomly sampled 50 data points from the automatic evaluation of discourse quality, with 10 data points for each rubric item, \ie ~\textit{Novelty}, \textit{Intent Alignment}, \textit{No Repetition} for question-asking utterances, and \textit{Consistency} and \textit{Engagement} for question-answering utterances, as defined in \refsec{sec:eval_metrics}. Each data point represents the automatic grading of one utterance on one rubric item. Two independent evaluators provided human grading. We calculate the Pearson correlation between the automatic evaluation scores and the average human grading scores. \reftab{tab:auto_grading_correlation} shows that the automatic rubric grading exhibits a positive correlation with human grading, with statistical significance observed for 4 out of the 5 rubric items. Additionally, the experimental results from the human evaluation with real users (Table \reftab{table:human_eval} and \reffig{fig:human_eval_pref}) also reveal similar findings to the automatic evaluation results, verifying our automatic evaluation setup.

\section{Human Evaluation Details}
\label{appendix:human_eval}

\begin{figure}[t]
    \centering 
    \resizebox{\columnwidth}{!}{%
    \includegraphics{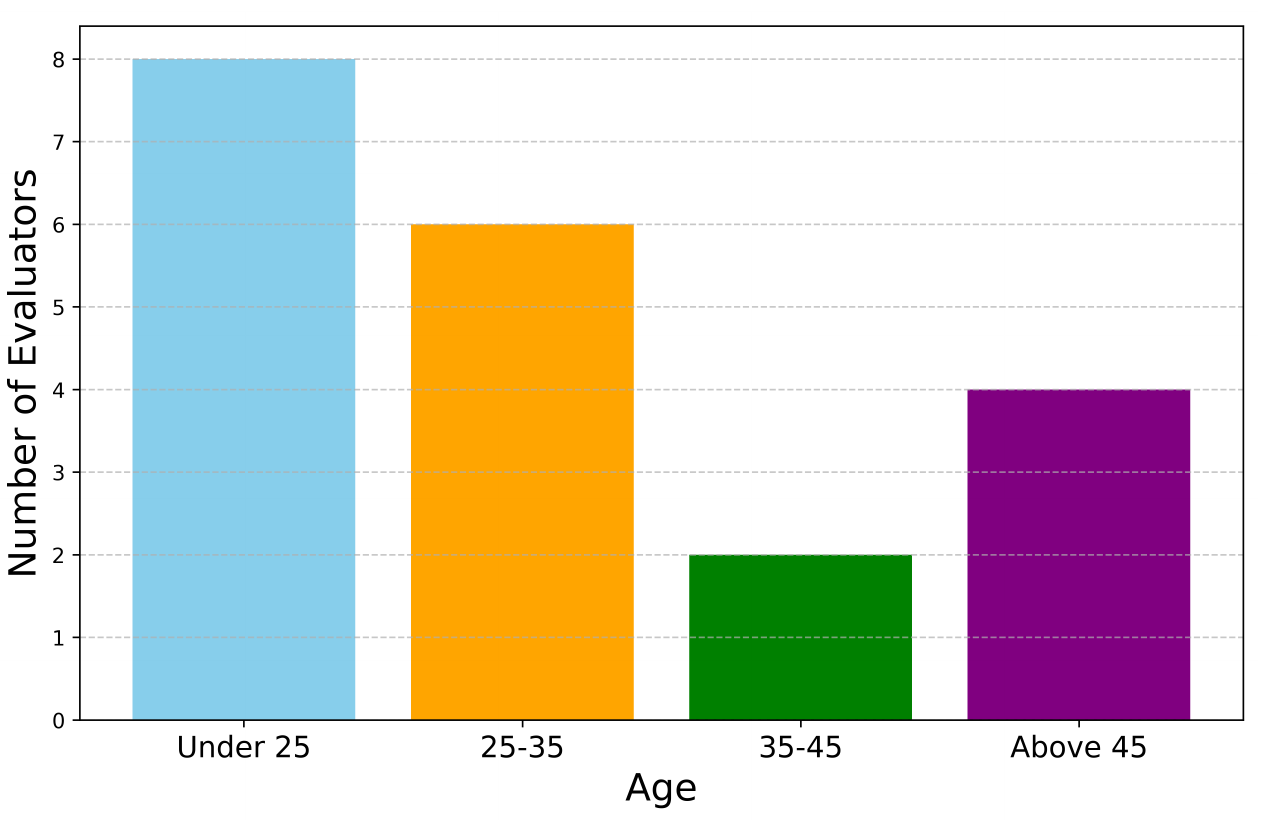}
    }
    \caption{Age distribution of participants in the human evaluation.}
    \label{fig:human_eval_age}
\end{figure}

\begin{figure}[t]
    \centering 
    \resizebox{\columnwidth}{!}{%
    \includegraphics{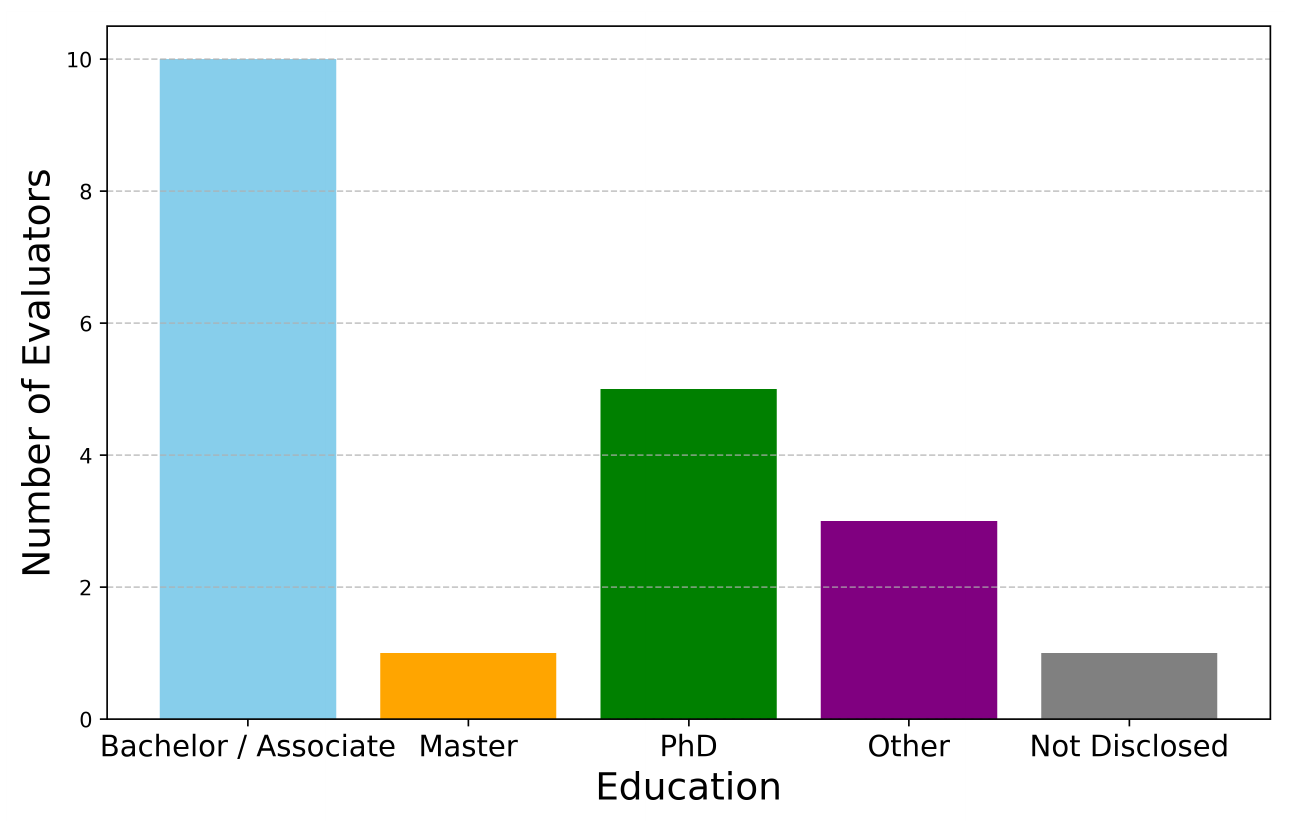}
    }
    \caption{Education level distribution of participants in the human evaluation.}
    \label{fig:human_eval_edu}
\end{figure}

\begin{table}
\centering
\resizebox{0.9\columnwidth}{!}{%
    
    \begin{tabular}{lc}
    \toprule
     & \textbf{Pearson Correlation} ($p$-value) \\
     \midrule
    Novelty  & 0.32 ($<$ 4e-1) \\
    Intent Alignment  & 0.55 ($<$ 2e-2) \\
    No Repetition  & 0.50 ($<$ 7e-3) \\
    Consistency  & 0.50 ($<$ 2e-3) \\
    Engagement  & 0.34 ($<$ 2e-2) \\
    \bottomrule
    \end{tabular}
}
    \caption{Pearson correlation between average human rubric grading scores and automatic rubric grading scores on discourse turn quality (n=50).}
    \label{tab:auto_grading_correlation}
\vspace{-1em}
\end{table}

\begin{figure*}[t]
    \centering 
    \resizebox{0.8\textwidth}{!}{%
    \includegraphics{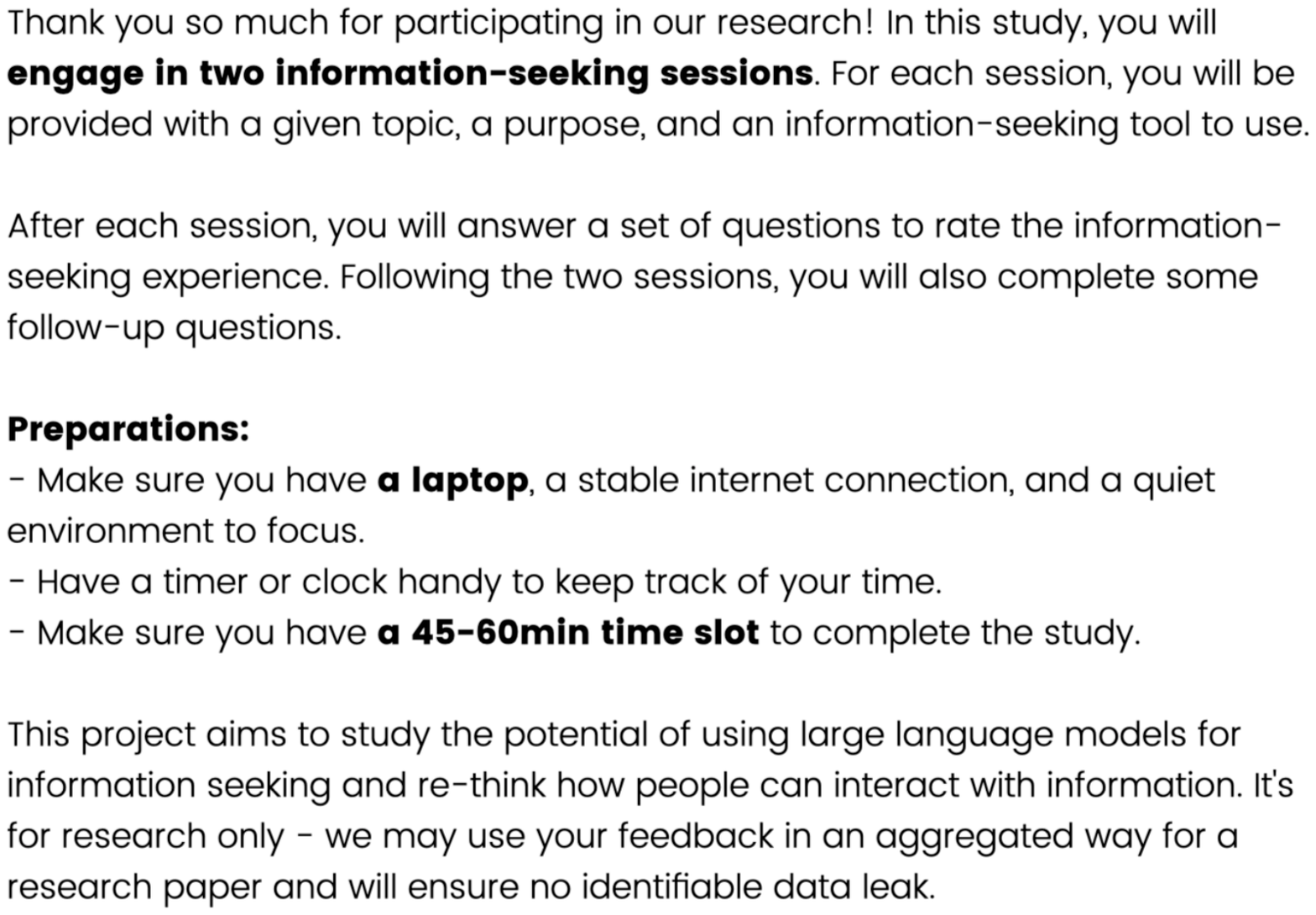}
    }
    \caption{Screenshot to get consent from participants for gathering feedback data during human evaluations using Qualtrics.}
    \label{fig:data_privacy_consent}
\end{figure*}

Human evaluation participants voluntarily provided demographic data, including their ages and highest education levels. As shown in~\reffig{fig:human_eval_age} and~\reffig{fig:human_eval_edu}), our human evaluation covers a diverse demographic. All participants gave consent to feedback data collection and we ensured no personal identifiable information was stored (see ~\reffig{fig:data_privacy_consent}). Feedback was collected via an online questionnaire platform \footnote{\url{https://www.qualtrics.com}} and a web application we built.

\begin{figure*}[t!]
    \centering 
    \resizebox{0.95\textwidth}{!}{%
    \includegraphics{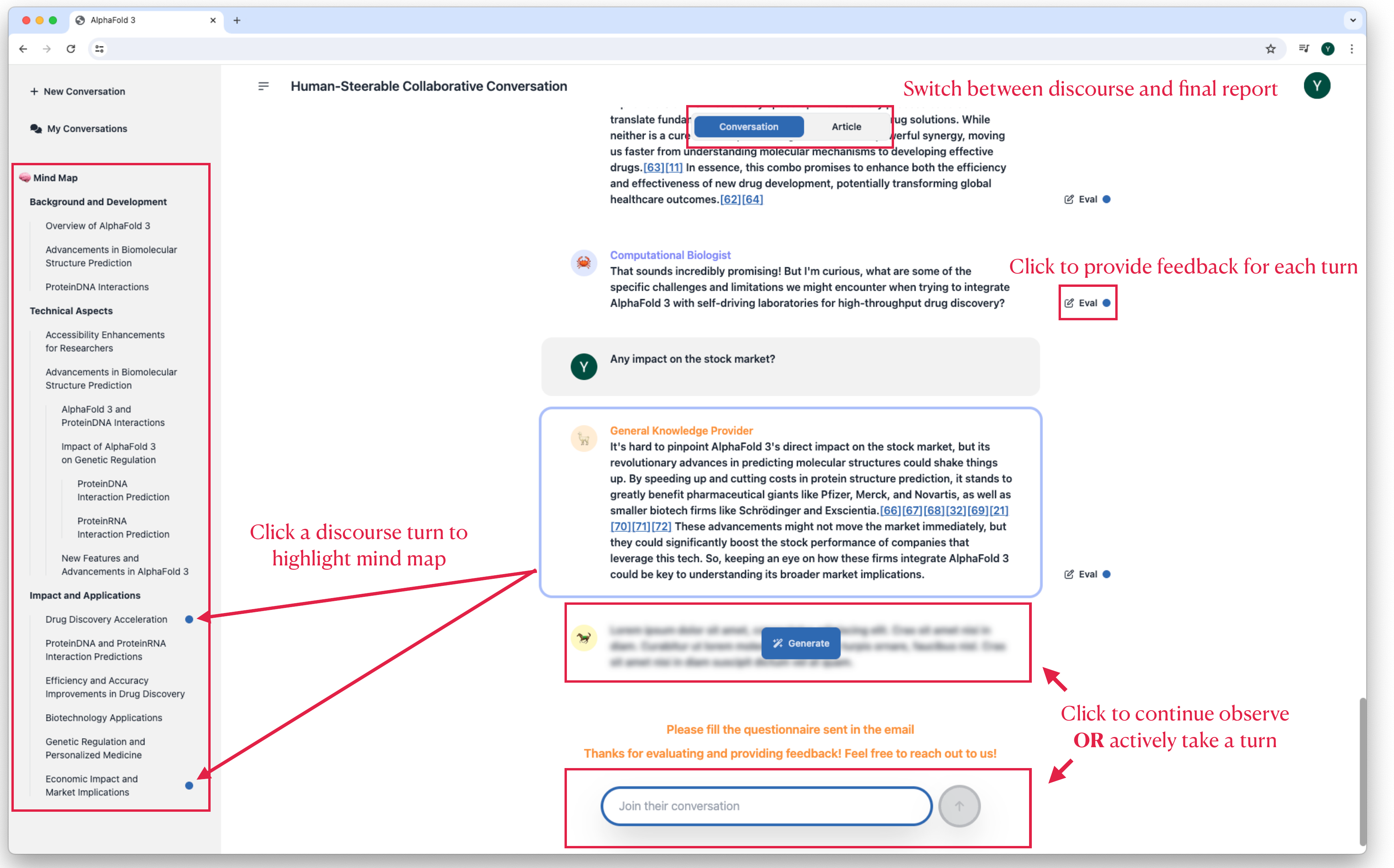}
    }
    \caption{Screenshot of the interactive web application for human evaluation.}
    \label{Fig.human_eval_demo}
\vspace{-0.5em}
\end{figure*}

The web application provides participants an interface to perform real-time interaction with \system. The web application has two modes, \system mode and RAG chatbot mode. \reffig{Fig.human_eval_demo} shows a screenshot of the web application in \system mode. The RAG chatbot mode is similar to the common chatbot interface.

\begin{figure*}[t]
    \centering 
    \resizebox{0.8\textwidth}{!}{%
    \includegraphics{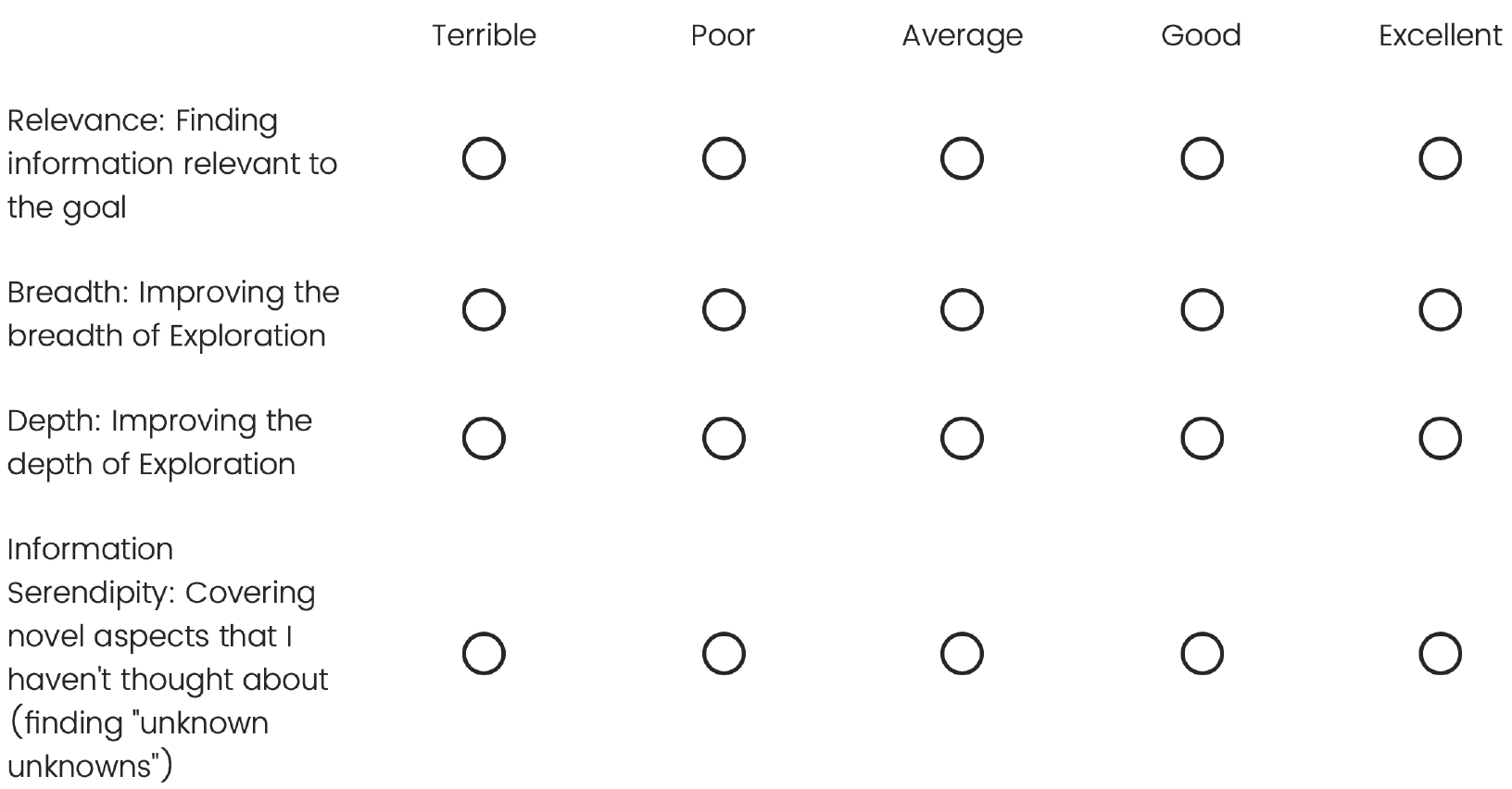}
    }
    \caption{Human evaluation grading rubrics for each method (search engine, RAG Chatbot, and \system). Evaluation results are shown in \reftab{table:human_eval}}.
    \label{fig:human_eval_method_rubric}
\end{figure*}

\begin{figure*}[t]
    \centering 
    \resizebox{0.8\textwidth}{!}{%
    \includegraphics{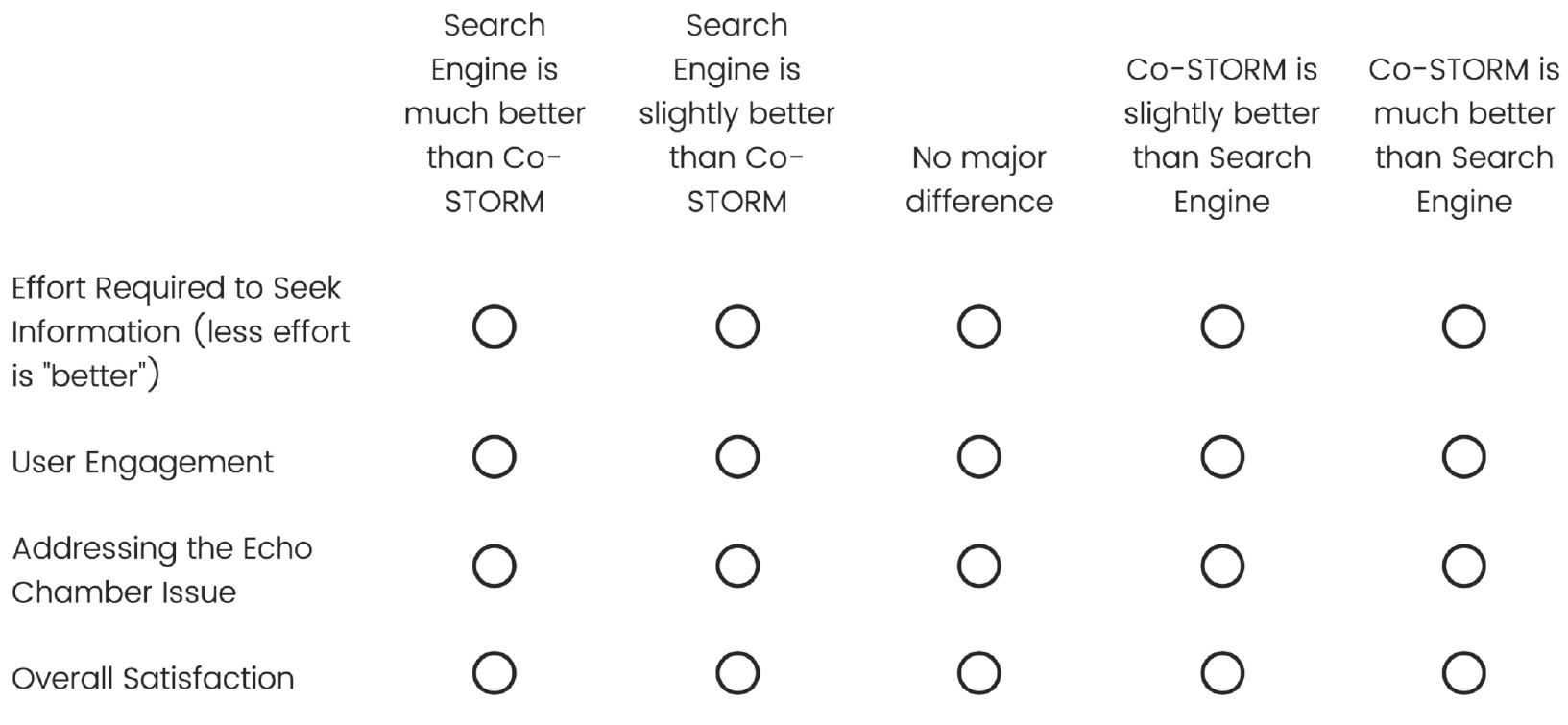}
    }
    \caption{Likert question for comparing \system with Google Search in human evaluation. Evaluation results are shown in~\reffig{fig:human_eval_pref}.}
    \label{fig:human_eval_pref_rubric}
\end{figure*}

\begin{figure*}[t]
    \centering 
    \resizebox{0.8\textwidth}{!}{%
    \includegraphics{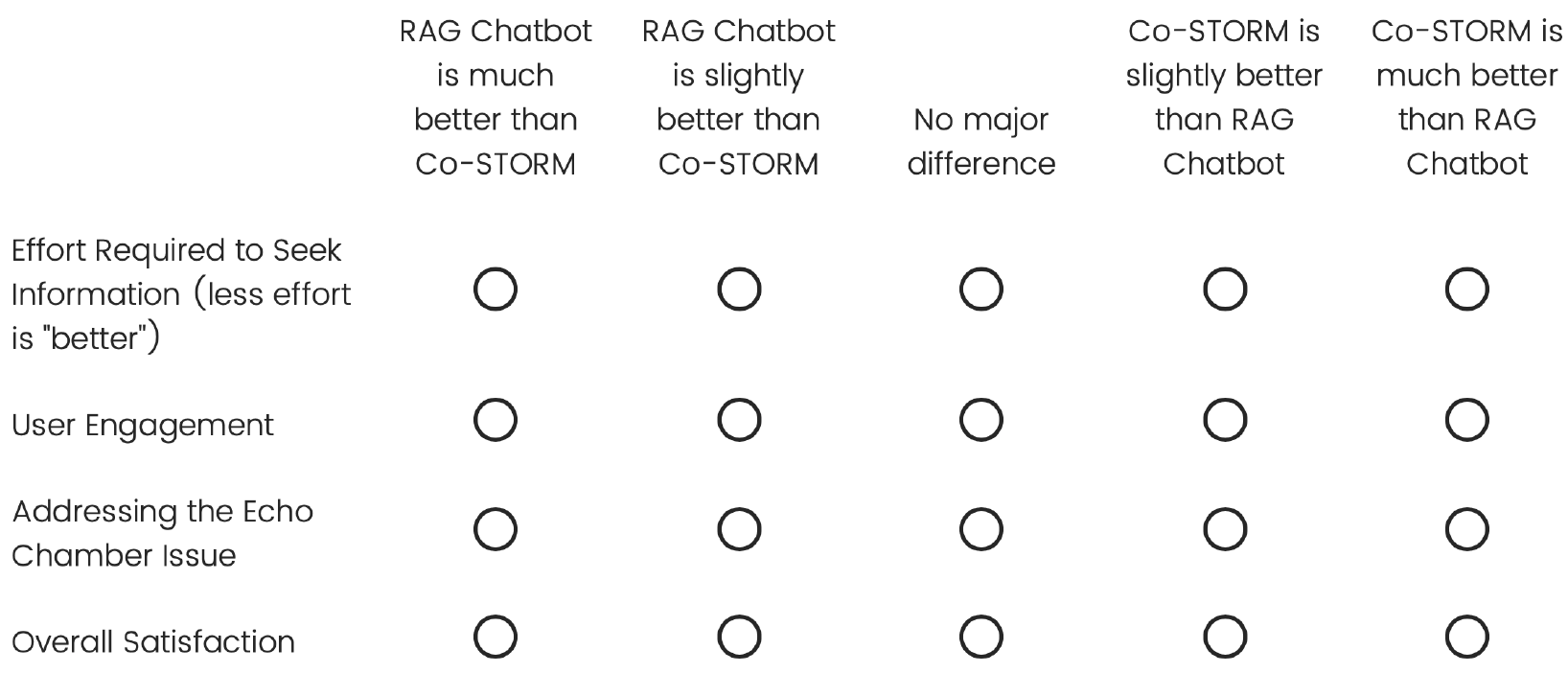}
    }
    \caption{Likert question for comparing \system with RAG Chatbot in human evaluation. Evaluation results are shown in~\reffig{fig:human_eval_pref}.}
    \label{fig:human_eval_pref_rag_rubric}
\end{figure*}

As discussed in \refsec{sec:human_eval}, we crafted five pairs of complex information-seeking tasks for human evaluation (see \reftab{table:human_eval_tasks}). After completing each task, participants were instructed to rate the information-seeking assistance system they used (\ie, Google Search, RAG Chatbot, or \system) from four grading aspects defined in \refsec{sec:eval_metrics} using 1 to 5 Likert scale (Likert question shown in \reffig{fig:human_eval_method_rubric}). After completing both two tasks, participants were asked to provide a pairwise preference by comparing \system with either Google Search (see \reffig{fig:human_eval_pref_rubric}) or the RAG chatbot (see \reffig{fig:human_eval_pref_rag_rubric}) with the Likert questions.

\section{Case Study}
We present two examples from different topics where the moderator effectively steers the conversation toward engaging directions. Example~\ref{showcase:showcase1} shows an example of discourse on the topic ``The effects of NMN supplements on human anti-aging'' where the \moderator effectively steers the ongoing discourse to the anti-aging benefits of personalized NMN and then further directs the discourse towards genetic profiling for personalized NMN supplementation plans. Example~\ref{showcase:showcase2} highlights \moderator effectively raises a new concept and shifts the discussion to the topic ``The Emergence of Artificial Super Intelligence: Future Prospects and Impacts''. The \moderator steers the ongoing discourse from technology hurdles, the role of computation power, societal impact, risk, and mitigation toward a discussion on the quantum digital twin.

Additionally, we include a complete discourse transcript (Appendix~\refsec{sec:alphafold_discourse}) and the associated report (Appendix~\refsec{sec:alphafold_report}) on the topic of “AlphaFold 3,” as referenced in \reffig{Fig.method}. In the discourse, the system initiates the discussion with steering by the \moderator, focusing on the background and development of AlphaFold 3, as well as the technical advancements in biomolecular structure prediction, protein-DNA interactions, and its impact on genetic regulation. The user then directs the discourse towards its applications. Several participants provide insights into AlphaFold 3’s applications in drug discovery, personalized medicine, and biotechnology. This is followed by a discussion on self-driving laboratories (SDLs), again steered by the \moderator. Finally, the user shifts the discussion towards the economic impact and market implications of AlphaFold 3.

\begin{table*}[!h]
\centering
\resizebox{\textwidth}{!}{%
\begin{tabular}{cp{0.95\linewidth}}
\toprule
\textbf{Topic} & \textbf{Goal} \\ 
\midrule
GPT-4 Omni &  \multirow{2}{*}{\parbox{0.95\linewidth}{To investigate the latest technology breakthrough and discover a unique angle to report on it, ensuring more people know about the technology.}} \\
AlphaFold 3  & \\
\midrule
Gaza war protests in US colleges &  \multirow{2}{*}{\parbox{0.8\linewidth}{To investigate the latest news and provide comprehensive coverage, ensuring people receive diverse perspectives on the events.}} \\
The conviction of Donald J. Trump in 2024  & \\
\midrule
Privacy Norm with Digital Technologies &  \multirow{2}{*}{\parbox{0.8\linewidth}{To gain an in-depth understanding of the topic and prepare for a one-hour presentation in a college reading group.}} \\
Copyright Issues with Language Models  & \\
\midrule
Social Organism &  \multirow{2}{*}{\parbox{0.8\linewidth}{To conduct a literature review on a given topic in preparation for a class discussion in a sociology course.}} \\
Social Statics and Social Dynamics  & \\
\midrule
China's dropping population in recent years &  \multirow{2}{*}{\parbox{0.8\linewidth}{To investigate the latest news and find an engaging angle to report it, incorporating background stories and connections to related events to enhance its appeal.}} \\
The Humanitarian crisis in Gaza in recent years  & \\
\bottomrule
\end{tabular}
}
\caption{Information-seeking tasks used in human evaluation.}
\label{table:human_eval_tasks}
\vspace{-0.3em}
\end{table*}



\onecolumn
\begin{figure*}
\input{prompts/mind_map_prompt}
\captionof{lstlisting}{Prompts used for dynamically updating the mind map in \system.}
\label{lst:mind_map_prompt}
\end{figure*}

\begin{figure*}
\input{prompts/expert_prompt}
\captionof{lstlisting}{Prompts used for simulating perspective-guided \experts in \system.}
\label{lst:participants_prompt}
\end{figure*}

\begin{figure*}
\input{prompts/moderator_prompt}
  \captionof{lstlisting}{Prompts used for simulating the \moderator in \system}
   \label{lst:moderator_prompt}
\end{figure*}
\twocolumn

\begin{table*}
\centering
\resizebox{\textwidth}{!}{%
\begin{tabular}{ll} 
\toprule
Criteria Description & \textbf{Broad Coverage}: Does the article provide an in-depth exploration of the topic and have good coverage?                                                                                                               \\
Score 1 Description  & Severely lacking; offers little to no coverage of the topic's primary aspects, resulting in a very narrow perspective.                                                                                                                        \\
Score 2 Description  & Partial coverage; includes some of the topic's main aspects but misses others, resulting in an incomplete portrayal.                                                                                                                              \\
Score 3 Description  & Acceptable breadth; covers most main aspects, though it may stray into minor unnecessary details or overlook some relevant points.                                                                                             \\
Score 4 Description  & Good coverage; achieves broad coverage of the topic, hitting on all major points with minimal extraneous information.                                                                      \\
Score 5 Description  & Exemplary in breadth; delivers outstanding coverage, thoroughly detailing all crucial aspects of the topic without including irrelevant information.                                                                                \\ 
\midrule
Criteria Description & \textbf{Novelty}: Does the report cover novel aspects that relate to the user's initial intent but are not directly derived from it?                                                                                              \\
Score 1 Description  & Lacks novelty; the report strictly follows the user's initial intent with no additional insights.                                                                                                                                     \\
Score 2 Description  & Minimal novelty; includes few new aspects but they are not significantly related to the initial intent.                                                                                              \\
Score 3 Description  & Moderate novelty; introduces some new aspects that are somewhat related to the initial intent.                                                                                  \\
Score 4 Description  & Good novelty; covers several new aspects that enhance the understanding of the initial intent.                                                                                                                      \\
Score 5 Description  & Excellent novelty; introduces numerous new aspects that are highly relevant and significantly enrich the initial intent.                                                                             \\ 
\midrule
Criteria Description & \textbf{Relevance and Focus}: How effectively does the report maintain relevance and focus, given the dynamic nature of the discourse?                                                                  \\
Score 1 Description  & Very poor focus; discourse diverges significantly from the initial topic and intent with many irrelevant detours.                                                                                                \\
Score 2 Description  & Poor focus; some relevant information, but many sections diverge from the initial topic.                                                                     \\
Score 3 Description  & Moderate focus; mostly stays on topic with occasional digressions that still provide useful information.                                                                                                                                    \\
Score 4 Description  & Good focus; maintains relevance and focus throughout the discourse with minor divergences that add value.                                                                 \\
Score 5 Description  & Excellent focus; consistently relevant and focused discourse, even when exploring divergent but highly pertinent aspects.  \\ 
\midrule
Criteria Description & \textbf{Depth of Exploration}: How thoroughly does the report explore the initial topic and its related areas, reflecting the dynamic discourse?                                                                     \\
Score 1 Description  & Very superficial; provides only a basic overview with significant gaps in exploration.                                                                    \\
Score 2 Description  & Superficial; offers some detail but leaves many important aspects unexplored.                                                                    \\
Score 3 Description  & Moderate depth; covers key aspects but may lack detailed exploration in some areas.                                                   \\
Score 4 Description  & Good depth; explores most aspects in detail with minor gaps.                                                                   \\
Score 5 Description  & Excellent depth; thoroughly explores all relevant aspects with comprehensive detail, reflecting a deep and dynamic discourse.                               \\
\bottomrule
\end{tabular}
}
\caption{Report scoring rubrics on a 1-5 scale for the Prometheus model.}
\label{table:final_report_rubric}
\end{table*}

\begin{table*}
\centering
\resizebox{\textwidth}{!}{%
\begin{tabular}{ll} 
\toprule
Criteria Description & \textbf{Novelty}: \makecell[l]{Evaluates the extent to which the conversation turn introduces new and unexpected information that is relevant to the topic at hand. \\ High novelty indicates the conversation is providing fresh insights or perspectives that the user might not have considered, \\ thereby enriching the dialogue and enhancing the user's understanding of the subject.}                                                                                                              \\
Score 1 Description  & The turn fails to introduce any new or unexpected information, repeating known facts or irrelevant content.                                                                                                                        \\
Score 2 Description  & The turn introduces some new information, but it is mostly predictable or only slightly relevant.                                                                                                                              \\
Score 3 Description  & The turn provides moderately novel information that is relevant and somewhat unexpected.                                                                                          \\
Score 4 Description  & The turn introduces new and relevant information that is largely unexpected, sparking interest.                                                                \\
Score 5 Description  & The turn consistently introduces highly novel and relevant information that is completely unexpected, significantly enhancing the conversation.                                                                               \\ 
\midrule
Criteria Description & \textbf{Engaging}: \makecell[l]{Measures how interesting and captivating the conversation turn is. An engaging turn holds the user's attention and encourages them \\ to continue interacting. It often includes elements that are thought-provoking, entertaining, or particularly relevant to the user's interests. }                                                                                                      \\
Score 1 Description  & The turn is dull and uninteresting, likely causing the user to lose interest.                                                                                                                                     \\
Score 2 Description  & The turn has limited engagement, with occasional interesting points but generally fails to captivate the user.                                                                                        \\
Score 3 Description  & The turn is moderately engaging, holding the user's interest but lacking captivating elements.                                                                                 \\
Score 4 Description  & The turn is engaging and interesting, encouraging further interaction with minor lapses.                                                                                                                    \\
Score 5 Description  & The turn is highly engaging, consistently holding the user's interest and encouraging further interaction.                                                                             \\ 
\midrule
Criteria Description & \textbf{Consistency}: \makecell[l]{Assesses whether the conversation turn contradicts previous statements or established facts. Minimizing contradictionsis essential \\ for maintaining trust and coherence in the conversation.  A high score indicates that the turn is free from inconsistencies and \\ logically fits with the preceding dialogue.}                                                                 \\
Score 1 Description  & The turn frequently contradicts previous statements or established facts, causing confusion.                                                                                               \\
Score 2 Description  & The turn occasionally contradicts itself, with some inconsistencies present.                                                               \\
Score 3 Description  & The turn is mostly free of contradictions, with only minor inconsistencies that do not significantly impact coherence.                                                                                                                                    \\
Score 4 Description  & The turn is nearly free of contradictions, with only very rare and minor inconsistencies.                                                                \\
Score 5 Description  & The turn is entirely free of contradictions, maintaining perfect coherence and logical consistency.  \\ 
\bottomrule
\end{tabular}
}
\caption{Question-answering turn scoring rubrics on a 1-5 scale for the Prometheus model.}
\label{table:question_answering_rubric}
\end{table*}

\begin{table*}
\centering
\resizebox{\textwidth}{!}{%
\begin{tabular}{ll} 
\toprule
Criteria Description & \textbf{Intent Alignment}: \makecell[l]{Assesses how well the conversation turn aligns with the user's latent intent or goals. It measures the relevance and appropriateness of the \\ response in contributing towards the user's overall objectives. High intent alignment ensures that the conversation stays focused on the user's\\ needs and drives towards meaningful outcomes. }                                                                                                      \\
Score 1 Description  & The turn does not align with the user's latent intent or goals, and may confuse the conversation's purpose.                                                                                                                    \\
Score 2 Description  & The turn slightly aligns with the user's latent intent, but does not significantly contribute to the overall goals.                                                                                       \\
Score 3 Description  & The turn moderately aligns with the user's latent intent, contributing to the overall goals in a limited way.                                                                                \\
Score 4 Description  & The turn aligns well with the user's latent intent, contributing meaningfully to the overall goals.                                                                                                              \\
Score 5 Description  & The turn perfectly aligns with the user's latent intent, significantly driving the conversation towards the overall goals.                                                                             \\ 
\midrule
Criteria Description & \textbf{Repetition}: \makecell[l]{Looks at the degree to which the conversation turn repeats information that has already been provided. Lower scores indicate higher repetition, \\ which can detract from the value of the conversation by failing to introduce new content. Ideally, each turn should add new information or \\ perspectives to the dialogue.}                                                                 \\
Score 1 Description  & The turn repeats information already provided without adding any new value.                                                                                             \\
Score 2 Description  & The turn has noticeable repetition, with limited new information added.                                             \\
Score 3 Description  & The turn includes some repetition, but provides enough new information to be moderately valuable.                                                                                                                                 \\
Score 4 Description  & The turn has minimal repetition, mostly introducing new and relevant information.                                                            \\
Score 5 Description  & The turn does not repeat any information, consistently providing new and valuable content.  \\ 
\bottomrule
\end{tabular}
}
\caption{Question-asking turn scoring rubrics on a 1-5 scale for the Prometheus model.}
\label{table:question_asking_rubric}
\end{table*}

\begin{figure*}
    \input{examples/discourse_snippet1}
    \caption{Example discourse highlight on the topic ``The effects of NMN supplements on human anti-aging''.}
    \label{showcase:showcase1}
\end{figure*}

\begin{figure*}
\input{examples/discourse_snippet2}
\caption{Example discourse highlight on the topic ``The Emergence of Artificial Super Intelligence: Future Prospects and Impacts''.}
    \label{showcase:showcase2}
\end{figure*}


\newpage
\onecolumn

\section{Full Collaborative Discourse Example}
\label{sec:alphafold_discourse}
\begin{showcase}[title=Topic: AlphaFold 3]
|\expertbg{Default Background Researcher:}|
Based on the available information, I cannot fully address the question in a conversational style. However, I can provide a detailed and accurate explanation in an informative manner.

AlphaFold 3 is the latest advancement in the field of molecular structure prediction, introduced in 2024 by Google DeepMind and Isomorphic Labs, both subsidiaries of Alphabet[2]. This groundbreaking AI model represents a substantial leap beyond its predecessor, AlphaFold 2, by expanding its capabilities to predict the structures and interactions of a broader spectrum of biomolecules[1][2]. 

Unlike AlphaFold 2, which primarily focused on single-chain proteins, AlphaFold 3 is designed to predict the structures of complex biomolecular systems. These systems include not only proteins but also DNA, RNA, post-translational modifications, and small molecules such as drugs and ligands[2][6][9]. The model operates by allowing scientists to input descriptions of these biomolecular complexes, which it then processes using a sophisticated neural network architecture. This architecture employs a custom Transformer with triangular attention and a diffusion process to generate the 3D coordinates of every atom in the specified system[6].

One of the most notable advancements in AlphaFold 3 is its potential application in drug design. By accurately predicting the structure and interactions of protein-ligand complexes, the model can aid in identifying and designing new molecules that could serve as effective drugs[3][4][10]. This has significant implications for the pharmaceutical industry, where Isomorphic Labs is already collaborating with companies to tackle real-world drug design challenges[3][4].

Researchers and scientists can access the majority of AlphaFold 3's capabilities for free through the newly launched AlphaFold Server, which is an easy-to-use research tool designed to make these advanced predictions accessible to a broader scientific community[3][4][7]. The introduction of AlphaFold 3 is expected to dramatically accelerate biological research by providing detailed atomic-level views of biomolecular interactions, thus enhancing our understanding of the biological world and facilitating the development of new treatments[9][11].

Overall, AlphaFold 3's ability to predict the structures of a wide variety of biomolecules more broadly and accurately than previous models makes it a revolutionary tool in the fields of structural biology and drug discovery[5][8].

|\expertbg{Bioinformatics Specialist:}|
Based on the available information, I cannot fully address the question. However, I can provide some insights into how AlphaFold 3 has improved accessibility and usability for researchers aiming to integrate its predictions into their biological studies.

AlphaFold 3 has introduced several significant enhancements over its predecessors, specifically designed to improve both accessibility and usability for researchers. One of the most notable advancements is the launch of the AlphaFold Server, which is a free, user-friendly platform that allows biologists to generate molecular complexes with just a few clicks[13][14]. This means that researchers, regardless of their computational resources or expertise in machine learning, can now easily access high-accuracy protein structure predictions. 

Additionally, AlphaFold 3 extends its predictive capabilities beyond just protein structures to include a variety of biomolecular systems, such as nucleic acids, small molecules, ions, and modified residues[12][17]. This broader scope of prediction is particularly beneficial for researchers involved in complex biological studies and drug discovery, as it allows them to model interactions across different types of molecules more accurately and comprehensively.

The new diffusion-based architecture of AlphaFold 3 enhances its prediction accuracy and reliability[12][16]. For example, the model shows a significant improvement in predicting protein-protein complexes and antibody-protein interfaces[18]. Such improvements are crucial for researchers focusing on immunology and therapeutic antibody development.

Despite these advancements, it's important to note that the downloadable code of AlphaFold 3 has not been released by DeepMind, unlike its previous models[14][15]. This may limit the flexibility for researchers who wish to modify or integrate the model deeply into their own systems. Nevertheless, the open accessibility of the AlphaFold Server provides a substantial step towards democratizing access to cutting-edge protein structure prediction technology.

In summary, AlphaFold 3 enhances accessibility through its user-friendly AlphaFold Server, allowing researchers to easily generate and use high-accuracy predictions. Its improved predictive capabilities across a range of biomolecular systems further support its integration into various biological studies. However, the absence of a downloadable code version may be a limitation for some users.

|\expertbg{Pharmaceutical Industry Expert:}|
Based on the available information, I cannot fully address the question. However, I can provide some relevant insights into the impact of AlphaFold 3 on drug discovery projects.

AlphaFold 3, developed by Google DeepMind and Isomorphic Labs, represents a significant advancement in the field of protein structure prediction and has transformative potential for drug discovery[21]. Compared to previous computational methods, AlphaFold 3 has improved the efficiency and accuracy of these projects in several ways.

Firstly, AlphaFold 3 has enhanced the prediction accuracy for protein structures, including complex interactions like antibody-antigen bindings, which are crucial for immunology and therapeutic antibody development[24]. This increased accuracy allows researchers to gain a better understanding of molecular interactions, which is essential for designing effective drugs.

Moreover, AlphaFold 3's ability to predict structures for nearly all molecules in the Protein Data Bank significantly broadens the scope of potential drug targets[20]. This capability is particularly valuable for tackling neglected diseases, which often suffer from a lack of research funding[25].

A notable example of AlphaFold 3's impact can be seen in a project conducted by Insilico and the University of Toronto. They applied AlphaFold to an end-to-end AI-powered drug discovery platform and successfully identified a new drug for a novel target to treat hepatocellular carcinoma (HCC)[23]. This project demonstrates how AlphaFold 3 can streamline the drug discovery process by integrating structure prediction with other AI models to identify effective therapeutic compounds efficiently.

Another example involves the discovery of a potent hit molecule, ISM042-2-048, using AlphaFold in combination with other AI tools. This molecule showed significant inhibitory activity and selective anti-proliferation effects, showcasing the practical application of AlphaFold in the hit identification phase of drug discovery[22].

In summary, AlphaFold 3 has made significant strides in improving the efficiency and accuracy of drug discovery projects. By providing highly accurate protein structure predictions and enabling the identification of new drug targets, it has the potential to revolutionize the development of new therapies. However, while the implications are promising, the full transformative impact on drug discovery may still be incremental at this stage[19].

|\expertbg{Structural Biologist:}|
Based on the available information, I cannot fully address the question. However, I can provide the most relevant answer based on the gathered information, and explain any limitations or gaps.

AlphaFold 3 introduces several specific advancements in predicting complex biomolecular structures, which significantly enhance our understanding of interactions involving proteins, DNA, RNA, and ligands at the atomic level. One of the key innovations in AlphaFold 3 is the updated diffusion-based architecture. This architecture enables joint structure prediction for complexes that include not only proteins but also nucleic acids (DNA and RNA), small molecules, ions, and modified residues[8]. This comprehensive capability allows for more accurate and detailed modeling of biomolecular interactions, surpassing many previous specialized tools[8][16].

Furthermore, AlphaFold 3 achieves unprecedented accuracy in predicting protein-ligand interactions, which are crucial for drug discovery. It is reported to be 50\% more accurate than the best traditional methods on the PoseBusters benchmark, and it does so without the need for any structural information input[26]. This makes AlphaFold 3 the first AI system to outperform physics-based tools in this domain[26].

In addition to protein-ligand interactions, AlphaFold 3 is capable of modeling the interactions between proteins and nucleic acids such as DNA and RNA[27]. This capability is particularly important for understanding the molecular mechanisms underlying genetic regulation and expression, as well as the role of RNA in various cellular processes.

AlphaFold 3's advancements also extend to the modeling of protein interactions with antibodies, which is another critical area for drug development[28]. The ability to predict these interactions with high accuracy can aid in the design of new therapeutic molecules and improve the efficiency of drug discovery pipelines[28].

Overall, AlphaFold 3 represents a significant leap forward in biomolecular structure prediction. Its ability to model a wide range of molecular interactions with high accuracy enhances our understanding of the atomic-level details of these interactions, which is essential for both basic biological research and practical applications in drug discovery and development[9][11].

While the gathered information highlights many advancements of AlphaFold 3, it does not provide exhaustive details on all potential applications and specific improvements over its predecessors. Nonetheless, the advancements mentioned above underline the transformative impact of AlphaFold 3 in the field of structural biology.

|\moderatorbg{\moderator:}|
That's fascinating! Speaking of advancements, how do you think AlphaFold 3's improved accuracy in predicting protein-DNA and protein-RNA interactions might deepen our insights into genetic regulation and expression?

|\expertbg{General Knowledge Provider:}|
Absolutely, AlphaFold 3 (AF3) is a game-changer for understanding genetic regulation and expression. Its diffusion-based architecture for predicting raw atom coordinates marks a significant leap, allowing it to model protein-DNA and protein-RNA interactions with much higher accuracy[29][8]. Unlike its predecessor, AlphaFold 2, which focused on individual protein structures, AF3 can now tackle these complex interactions, making it possible to gain deeper insights into how genetic processes are regulated.

What's really impressive is the 50\% improvement in prediction accuracy over existing methods, with some categories even doubling in accuracy[30][33]. This means we can now more reliably model these interactions, shedding light on the mechanisms of gene regulation and expression at a molecular level.

Additionally, AF3's ability to predict interactions across all of life's molecules, including nucleic acids, supports a wide range of scientific applications-from drug design to genomics research[31][32]. For instance, understanding protein-RNA interactions is crucial for RNA-based therapeutics, opening new avenues in medical research. 

So, in a nutshell, AF3 is significantly enhancing our understanding of genetic regulation and expression by providing highly accurate models of these essential interactions.

|\expertbg{Biochemist/Structural Biologist:}|
Absolutely, the strides AlphaFold 3 (AF3) has made in predicting protein-DNA and protein-RNA interactions are nothing short of revolutionary for genetic research and personalized medicine. With its diffusion-based architecture, AF3 goes beyond what AlphaFold 2 could achieve by predicting raw atom coordinates, allowing it to model a wide range of biomolecular interactions with high accuracy, including those involving nucleic acids, small molecules, ions, and modified residues [29][8].

One of the standout features of AF3 is its accuracy; it's shown to improve prediction accuracy by at least 50\%, and in some cases, even double it [30]. This is a game-changer for understanding gene regulation and expression, providing atomic-level insights into how proteins interact with DNA and RNA [11][5]. Imagine being able to visualize these interactions in such detail-this could lead to breakthroughs in developmental biology, oncology, and neurogenetics [31].

Moreover, AF3's capabilities have profound implications for personalized medicine. By accurately modeling protein-RNA interactions, it helps us understand how genetic variations affect individual responses to treatments, paving the way for more personalized and effective therapies [36][37]. It's not just a tool for basic research but a potential catalyst for transforming medical practices and treatment strategies.

|\expertbg{Geneticist:}|
Building on AlphaFold 3's (AF3) advancements, its predictive capabilities can really push the boundaries of gene regulation research and genetic therapies. AF3 doesn't just improve on AlphaFold 2's accuracy; it broadens the scope to predict complexes involving proteins, nucleic acids, small molecules, and ions[8][41]. This expanded understanding of protein interactions in their full biological context, such as with DNA, RNA, and ligands, is crucial for effective gene regulation and therapy[38][42].

AF3's enhanced ability to model protein-ligand interactions with high precision can significantly aid in designing targeted genetic therapies and drugs[8]. For instance, its predecessor has already contributed to drug discovery, including cancer immunotherapy and COVID-19 vaccine development[43][40]. With AF3, we can expect even more precise modeling of these interactions, leading to the identification of novel therapeutic targets.

Moreover, DeepMind's commitment to open science, sharing over 200 million protein structures, accelerates collaborative efforts in genetic therapies by providing valuable data to researchers globally[39]. This open-access resource is a goldmine for identifying new protein targets and understanding their roles in genetic regulation, paving the way for breakthroughs in gene therapy and personalized medicine. 

In essence, AF3's advanced capabilities offer unprecedented opportunities in both scientific research and therapeutic development, enhancing our ability to identify novel protein targets for genetic therapies[38][8][41][42].

|\userbg{User:}|
What are AlphaFold3 impacts and what are its applications?

|\expertbg{General Knowledge Provider:}|
Sure, let's dive into what makes AlphaFold 3 such a game-changer and where it's making waves.

AlphaFold 3, developed by DeepMind, has taken our ability to predict protein structures to a whole new level, hitting almost 99\% accuracy for human proteins[44][45]. This isn't just a cool tech feat; it's a leap in how we understand biology, so much so that some think it could snag a Nobel Prize down the line[45].

For researchers, AlphaFold 3 is like a supercharged toolkit. It complements traditional methods like X-ray crystallography, cryo-EM, and NMR spectroscopy, speeding up the pace of discovery and offering new insights into the molecular basis of diseases[49]. 

One of the biggest wins is in drug discovery. By predicting protein interactions with other molecules, AlphaFold 3 can help identify new drug candidates much faster, tackling one of the biggest hurdles in the field[13][1].

Then there's the AlphaFold Database, now with over 200 million protein structure predictions. This open-access treasure trove is leveling the playing field, allowing researchers everywhere to tap into top-notch structural data[13][48]. It's fueling advancements in enzyme design, cancer research, and even vaccine development[46][47].

Beyond proteins, AlphaFold 3 is branching out to predict interactions involving DNA, RNA, and small molecules, opening doors in genomics, biotechnology, and synthetic biology[8][1].

Of course, it's not a silver bullet. Turning these insights into real-world applications will need lots of teamwork between scientists and industry experts[45].

All in all, AlphaFold 3 is revolutionizing structural biology and paving the way for breakthroughs in drug discovery, biotechnology, and beyond. Its impact is vast, touching everything from basic research to practical medical and industrial innovations.

|\expertbg{Biochemist:}|
Absolutely, AlphaFold 3 (AF3) represents a huge leap in protein structure prediction, especially with its diffusion-based architecture. Unlike AlphaFold 2's Evoformer model, AF3 can accurately model interactions between proteins and various biomolecules like DNA, RNA, and small molecules with remarkable precision[52][29]. This is groundbreaking for genetic regulation research, as it allows us to predict how proteins interact with DNA and uncover the mechanisms behind gene activation and repression[50][8][53].

This enhanced capability is crucial for developing targeted gene therapies. With AF3, we can pinpoint key regulatory elements and design interventions to modify gene expression precisely. This has the potential to lead to innovative treatments for genetic disorders, where adjusting gene function is essential[9][31][34].

However, it's worth mentioning that access to AlphaFold 3 is currently limited by DeepMind, which could slow down its widespread application[51]. Despite this hurdle, the advancements made by AF3 offer a bright future for genetic research and therapeutic development.

In short, AlphaFold 3's ability to predict protein-DNA interactions with high accuracy could revolutionize our understanding of genetic regulation and pave the way for targeted gene therapies, even if access constraints may delay its broader use.

|\expertbg{Pharmaceutical Industry Representative:}|
Absolutely, the potential of AlphaFold 3 in drug discovery is immense. Given its advanced ability to predict 3D structures of complex biomolecular interactions-including proteins, DNA, RNA, and small molecules-it offers a significant advantage in understanding how drugs bind to their targets [55][7]. 

One of the standout features of AlphaFold 3 is its capability to predict the effects of post-translational modifications and ions on molecular systems [55]. This can be a game-changer for identifying new drug targets that were previously considered too challenging to tackle [55][56]. 

By improving the accuracy of drug metabolism and toxicity predictions, AlphaFold 3 can substantially cut down the number of necessary wet lab experiments, saving both time and money [58]. Moreover, its near-complete prediction of human protein structures opens up unprecedented opportunities for fundamental research and therapeutic development [44]. 

While these advancements are promising, some experts still see the current impact on drug discovery as incremental, requiring further validation [54]. Nonetheless, the potential for AlphaFold 3 to streamline drug discovery, especially for complex diseases, is incredibly exciting [21][20].

|\moderatorbg{\moderator:}|
That's a great point. How do you think integrating AlphaFold 3 with emerging technologies like self-driving laboratories could further accelerate and innovate the drug discovery process? (inspired by [60][61])

|\expertbg{General Knowledge Provider:}|
Absolutely, integrating AlphaFold 3 with self-driving laboratories (SDLs) can truly revolutionize drug discovery. AlphaFold 3's prowess in predicting 3D structures of complex biomolecular systems offers deep insights into molecular interactions crucial for drug design[55][9]. This means we can now target proteins that were previously considered untouchable with remarkable precision[55][44].

On the other hand, SDLs bring in the power of machine learning, lab automation, and robotics to conduct high-throughput experiments autonomously[64][65]. By merging AlphaFold 3's predictive capabilities with SDLs, we can streamline the process of identifying and optimizing drug candidates. SDLs can swiftly test numerous hypotheses and chemical compounds, guided by the insights from AlphaFold 3, making the identification of promising therapeutic candidates more efficient than ever[64].

Together, AlphaFold 3 and SDLs not only speed up the discovery process but also translate fundamental scientific insights into practical drug solutions. While neither is a cure-all alone, their integration creates a powerful synergy, moving us faster from understanding molecular mechanisms to developing effective drugs[63][11].

In essence, this combo promises to enhance both the efficiency and effectiveness of new drug development, potentially transforming global healthcare outcomes[62][64].

|\expertbg{Computational Biologist:}|
That sounds incredibly promising! But I'm curious, what are some of the specific challenges and limitations we might encounter when trying to integrate AlphaFold 3 with self-driving laboratories for high-throughput drug discovery?

|\userbg{User:}|
Any impact on the stock market?

|\expertbg{General Knowledge Provider:}|
It's hard to pinpoint AlphaFold 3's direct impact on the stock market, but its revolutionary advances in predicting molecular structures could shake things up. By speeding up and cutting costs in protein structure prediction, it stands to greatly benefit pharmaceutical giants like Pfizer, Merck, and Novartis, as well as smaller biotech firms like Schrodinger and Exscientia[66][67][68][32][69][21][70][71][72]. These advancements might not move the market immediately, but they could significantly boost the stock performance of companies that leverage this tech. So, keeping an eye on how these firms integrate AlphaFold 3 could be key to understanding its broader market implications.

|\expertbg{Financial Analyst:}|
Absolutely, let's dive into how AlphaFold 3 is being integrated into the workflows of both pharmaceutical giants and smaller biotech firms, and the financial implications of this integration.

AlphaFold 3, developed by Google DeepMind, goes beyond predicting individual protein structures to modeling interactions between multiple proteins, DNA, RNA, and small molecules[73][74]. This makes it a game-changer in drug discovery and development.

Big pharma companies, especially those partnering with Isomorphic Labs, are using AlphaFold 3 to tackle complex drug design challenges[80][81]. By integrating this tech, they can speed up the drug discovery process, cut costs, and improve prediction accuracy, which enhances their development pipelines[78][79].

Smaller biotech firms are also leveraging AlphaFold 3 through platforms like Benchling, which incorporates AlphaFold beta for predicting 3D structures of novel proteins[75][77]. This helps them innovate more efficiently without needing extensive computational resources or deep machine learning expertise[78].

However, scaling AlphaFold 3 poses challenges like building the necessary IT infrastructure and ensuring data integrity and interoperability[76].

In essence, both large and small firms are using AlphaFold 3 to boost efficiency and innovation in drug discovery, with potentially significant long-term financial benefits.
\end{showcase}

\newpage
\section{Final Report Example}
\label{sec:alphafold_report}
\renewcommand{\arraystretch}{1.5} 
\begin{longtable}{p{\textwidth}}
\toprule
\texttt{\small{\textbf{AlphaFold 3}}} \\
\texttt{\small{\textbf{ \# Background and Development }}} \\
\vspace{1pt}
\texttt{\small{\textbf{ \#\# Overview of AlphaFold 3 }}} \\
\texttt{\small{ AlphaFold 3, co-developed by Google DeepMind and Isomorphic Labs, was announced on May 8, 2024, and represents a significant advancement in the prediction of biomolecular structures and interactions[2][7]. Unlike its predecessor, AlphaFold 2, AlphaFold 3 extends its capabilities beyond single-chain proteins to a wide array of biomolecules including DNA, RNA, post-translational modifications, and small molecules such as ligands, which are critical in drug discovery[1][2]. }} \\
\texttt{\small{ The AI model employs a neural network architecture that utilizes a custom Transformer with triangular attention and a diffusion process to generate 3D coordinates of every atom within the specified biomolecular system[6]. This allows researchers to input descriptions of complex biomolecular systems and receive highly accurate predictions of their three-dimensional structures[6]. }} \\
\texttt{\small{ One of the standout features of AlphaFold 3 is its ability to predict the structures of protein complexes with enhanced accuracy, particularly in protein-DNA and protein-RNA interactions. This improvement is expected to facilitate groundbreaking discoveries in gene regulation and expression, potentially revolutionizing genetic research and personalized medicine[5]. }} \\
\texttt{\small{ To facilitate broader scientific research, the capabilities of AlphaFold 3 are accessible for free through the AlphaFold Server, an easy-to-use research tool[3][4]. This accessibility is aimed at accelerating biological research and drug discovery processes by enabling more precise identification of drug targets and reducing the time and costs associated with developing new medications, especially for complex diseases[7][10]. Isomorphic Labs is also collaborating with pharmaceutical companies to leverage AlphaFold 3's potential in real-world drug design challenges, with the ultimate goal of developing new life-changing treatments for patients[3][4]. }} \\
\vspace{1pt}
\texttt{\small{\textbf{ \#\# Advancements in Biomolecular Structure Prediction }}} \\
\texttt{\small{ AlphaFold 3 represents a significant leap in the field of biomolecular structure prediction, building upon the successes of its predecessor, AlphaFold 2. The new model boasts a substantially updated diffusion-based architecture, which enables joint structure prediction of not only proteins but also nucleic acids, small molecules, ions, and modified residues[8][38]. This expansion beyond proteins marks a pivotal advancement, allowing for a more comprehensive understanding of complex biomolecular interactions and enhancing the accuracy of these predictions. }} \\
\texttt{\small{ One of the most notable achievements of AlphaFold 3 is its improved accuracy, which far surpasses many specialized tools previously used for protein-ligand interactions[8]. This heightened precision has broad implications, particularly in identifying novel protein targets for genetic therapies, thereby offering unprecedented opportunities in gene regulation research[38][39]. Furthermore, the model's capabilities extend to predicting the structures of complexes involving DNA and RNA, which could significantly advance our understanding of genetic regulation and aid in the development of targeted gene therapies[8][42]. }} \\
\texttt{\small{ DeepMind's commitment to open science has also played a crucial role in the impact of AlphaFold 3. By freely publishing the predicted structures of over 200 million proteins, DeepMind fosters collaboration and knowledge sharing, accelerating scientific discoveries and pharmaceutical development worldwide[39]. This initiative ensures that the benefits of AlphaFold 3 are widely accessible, contributing to collective scientific progress. }} \\
\texttt{\small{ The applications of AlphaFold 3 extend beyond academic research, impacting practical fields such as vaccine development. For instance, the structure-guided design of COVID-19 vaccines by companies like Pfizer, Moderna, and Johnson \& Johnson benefited from advancements in protein structure prediction, highlighting the model's potential in addressing global health challenges[40]. }} \\
\texttt{\small{ Despite its significant advancements, some researchers have noted limitations in AlphaFold 3's accuracy for a subset of its predictions, and the model does not fully reveal the underlying mechanisms of protein folding[42]. Nevertheless, the broader understanding of biomolecular contexts provided by AlphaFold 3, including the interactions of drug targets with protein binding partners, DNA, RNA, and ligand cofactors, is expected to lead to more effective therapeutic interventions[43]. This richer contextual insight underscores the potential for rational, structure-based drug design, as demonstrated in the examination of TIM-3, a potential target for cancer immunotherapy[43]. }} \\
\vspace{1pt}
\texttt{\small{\textbf{ \#\# ProteinDNA Interactions }}} \\
\texttt{\small{ AlphaFold 3 represents a significant advancement in the field of computational biology, specifically in the prediction of the structure of biomolecular systems. Building upon the foundational work of AlphaFold 2, which accurately predicted the structure of individual proteins, AlphaFold 3 extends these capabilities to include complex interactions involving multiple proteins, DNA, RNA, and small molecule ligands[9][11]. This includes an accurate atomic-level view of how these biomolecules come together and interact, providing critical insights into the structural impact of post-translational modifications and ions[11]. }} \\
\texttt{\small{ By providing detailed predictions of protein-DNA interactions, AlphaFold 3 enhances our understanding of gene regulation and the molecular basis of various diseases. This capability is crucial for advancing drug discovery, as it allows researchers to identify potential therapeutic targets more accurately and to understand how drugs can modify these interactions to produce desired effects[9][11]. The integration of AlphaFold 3 with emerging technologies, such as self-driving laboratories, promises to further accelerate and innovate the drug discovery process by automating the synthesis, testing, and optimization of new drug candidates based on precise structural data[9]. }} \\
\vspace{1pt}
\texttt{\small{\textbf{ \# Technical Aspects }}} \\
\vspace{1pt}
\texttt{\small{\textbf{ \#\# Accessibility Enhancements for Researchers }}} \\
\texttt{\small{ AlphaFold 3 significantly enhances accessibility and usability for researchers looking to integrate its predictions into their biological studies. One of the major strides in accessibility is the launch of the AlphaFold Server, a free and user-friendly research tool powered by AlphaFold 3. This server is touted as the most accurate tool globally for predicting protein interactions with other molecules within the cell. Researchers, irrespective of their computational resource availability or machine learning expertise, can generate molecular complexes with just a few clicks on a single platform[13]. }} \\
\texttt{\small{ Google DeepMind has made AlphaFold Server openly accessible to facilitate the global research community's use of AlphaFold 3, driving advancements in fields such as drug discovery, biotechnology, genomics, and our foundational understanding of biological systems. However, it is noteworthy that, unlike its predecessors, DeepMind has not released the downloadable code for AlphaFold 3[14]. Despite this, the open access to the AlphaFold Database, which houses over 200 million protein structure predictions, continues to accelerate scientific research[13]. }} \\
\texttt{\small{ Additionally, the structural predictions provided by AlphaFold 3 extend beyond proteins to include nucleic acids, small molecules, ions, and modified residues. This comprehensive predictive ability marks a substantial improvement over the specialized models of AlphaFold 2, which were more limited in scope[12][16][17]. The enhancements in prediction accuracy, particularly for protein-protein complexes and antibody-protein interfaces, offer researchers more reliable data to advance their studies[18]. These advancements collectively contribute to a more accessible and powerful tool for the scientific community. }} \\
\vspace{1pt}
\texttt{\small{\textbf{ \#\# Advancements in Biomolecular Structure Prediction }}} \\
\vspace{1pt}
\texttt{\small{\textbf{ \#\#\# AlphaFold 3 and ProteinDNA Interactions }}} \\
\texttt{\small{ AlphaFold 3 has revolutionized the modeling of protein-DNA interactions, an essential component in understanding genetic regulation. The updated diffusion-based architecture of AlphaFold 3 enables the joint structure prediction of complexes, including not just proteins, but also nucleic acids such as DNA, small molecules, ions, and modified residues[8]. This comprehensive approach allows for significantly improved accuracy over many previous specialized tools, especially in predicting protein-ligand interactions[8]. By accurately modeling these interactions, AlphaFold 3 provides deeper insights into the mechanisms of genetic regulation and opens new avenues for developing targeted gene therapies[8]. }} \\
\vspace{1pt}
\texttt{\small{\textbf{ \#\#\# Impact of AlphaFold 3 on Genetic Regulation }}} \\
\vspace{1pt}
\texttt{\small{\textbf{ \#\#\#\# ProteinDNA Interaction Prediction }}} \\
\texttt{\small{ AlphaFold 3 marks a significant advancement in the prediction of protein-DNA interactions, offering enhanced capabilities compared to its predecessor. Unlike AlphaFold 2, which was optimized for predicting the structure of individual proteins, AlphaFold 3 employs a diffusion-based model that predicts raw atom coordinates, allowing it to accurately model an array of biomolecular interactions including those between proteins and nucleic acids like DNA and RNA[29]. }} \\
\texttt{\small{ The shift to a diffusion-based architecture enables AlphaFold 3 to achieve a remarkable improvement in prediction accuracy. Specifically, the model shows at least a 50\% improvement in predicting the interactions of proteins with other molecule types, and in certain crucial categories, the accuracy has doubled compared to existing methods[30]. This enhanced prediction capability can lead to groundbreaking discoveries in gene regulation mechanisms and revolutionize our approach to genetic research and personalized medicine[30]. }} \\
\texttt{\small{ Introduced in collaboration with Isomorphic Labs, AlphaFold 3 goes beyond proteins to encompass a broad spectrum of biomolecules, including DNA, RNA, and small molecules known as ligands. This comprehensive approach opens new avenues for transformative science, from developing biorenewable materials and more resilient crops to accelerating drug design and genomics research[31]. By accurately predicting the interactions of proteins with DNA, AlphaFold 3 holds the potential to significantly advance our understanding of genetic regulation and assist in the development of targeted gene therapies[29][31]. }} \\
\vspace{1pt}
\texttt{\small{\textbf{ \#\#\#\# ProteinRNA Interaction Prediction }}} \\
\texttt{\small{ AlphaFold 3 has marked a significant leap forward in the field of structural biology by enhancing its prediction accuracy for protein-DNA and protein-RNA interactions. Building upon the foundational work of AlphaFold 2, the latest iteration of AlphaFold developed by Google's DeepMind and Isomorphic Labs in London can now predict the structure and interactions of a wide array of biomolecular systems with unprecedented precision[5][11][37]. This includes a dramatic improvement, with at least a 50\% enhancement in accuracy for interactions between proteins and other molecule types compared to existing methods, and in certain crucial categories, the prediction accuracy has doubled[33]. }} \\
\texttt{\small{ These advancements hold transformative potential for understanding genetic regulation and expression, as the more accurate predictions can provide deeper insights into the mechanisms of gene regulation[36][37]. Such detailed atomic-level views of molecular interactions are expected to revolutionize approaches in genetic research and personalized medicine, paving the way for groundbreaking discoveries in how genes are regulated and expressed within biological systems[5][11]. This progress also means that the model is not limited to proteins but extends to DNA, RNA, and other small molecules, enabling a more comprehensive understanding of biomolecular dynamics[11]. }} \\
\vspace{1pt}
\texttt{\small{\textbf{ \#\#\# New Features and Advancements in AlphaFold 3 }}} \\
\texttt{\small{ AlphaFold 3 introduces several groundbreaking features and advancements in the field of biomolecular structure prediction. One of the most significant improvements is the ability to predict the structure of a wide variety of biomolecular systems more broadly and accurately than its predecessor, AlphaFold 2. This has been achieved through the use of diffusion techniques to enhance the underlying architectural model, allowing for more general predictions[16]. }} \\
\texttt{\small{ Notably, AlphaFold 3 has set a new benchmark in accuracy for predicting drug-like interactions, including the binding of proteins with ligands and antibodies with their target proteins. It is 50\% more accurate than the best traditional methods on the PoseBusters benchmark, and it achieves this without requiring any input of structural information. This makes AlphaFold 3 the first AI system to outperform physics-based tools for biomolecular structure prediction[26]. }} \\
\texttt{\small{ Another significant advancement is AlphaFold 3's ability to model proteins interacting not only with other proteins but also with other biomolecules, such as DNA and RNA strands[27]. This capability is particularly valuable for understanding complex biological processes and interactions at the atomic level. Additionally, AlphaFold 3 excels in modeling protein-ligand interactions, a feature crucial for drug discovery efforts[27][28]. Accurate predictions of protein-ligand structures facilitate the identification and design of new molecules, which could potentially be developed into therapeutic drugs[28]. }} \\
\texttt{\small{ Early analyses have shown that AlphaFold 3 greatly outperforms AlphaFold 2.3 in certain protein structure prediction problems relevant to drug discovery, such as antibody binding[28]. This underscores the system's potential to significantly impact the pharmaceutical industry by improving the efficiency and accuracy of drug discovery processes[28]. }} \\
\vspace{1pt}
\texttt{\small{\textbf{ \# Impact and Applications }}} \\
\texttt{\small{ AlphaFold 3, developed by Google DeepMind in collaboration with Isomorphic Labs, has made significant strides in biotechnology by accurately predicting the structure and interactions of a wide range of biological molecules, including proteins, DNA, RNA, and small molecules such as drugs[1][7][9]. This advancement has substantial implications for several fields, most notably drug discovery and genetic research. }} \\
\texttt{\small{ One of the key impacts of AlphaFold 3 is its potential to dramatically accelerate the drug discovery process. By enabling precise identification of drug targets, it reduces both the time and costs associated with developing new medications, particularly for complex diseases[7][19][20][21]. The model's ability to predict how proteins interact with other molecules offers invaluable insights into the mechanisms of diseases and the development of targeted therapies[7][11]. Additionally, the integration of AlphaFold 3 with emerging technologies like self-driving laboratories could further innovate the drug discovery process, enhancing efficiency and accuracy[9][11]. }} \\
\texttt{\small{ In genetic research, AlphaFold 3's capability to predict protein-DNA interactions could significantly advance our understanding of genetic regulation, thereby aiding in the development of targeted gene therapies[8]. By providing an atomic-level view of biomolecular systems, including the structural impact of post-translational modifications and ions, AlphaFold 3 deepens our understanding of the biological world[11]. }} \\
\texttt{\small{ The introduction of AlphaFold Server, a free and accessible research tool powered by AlphaFold 3, has further democratized access to this groundbreaking technology. Researchers can now generate molecular complexes with minimal computational resources or expertise in machine learning, accelerating scientific research across the globe[13]. The server and the AlphaFold database provide open access to over 200 million protein structure predictions, fostering an environment of collaborative scientific discovery[13][20]. }} \\
\vspace{1pt}
\texttt{\small{\textbf{ \#\# Drug Discovery Acceleration }}} \\
\texttt{\small{ AlphaFold 3 represents a significant advancement in drug discovery, offering the potential to revolutionize the field by enabling more precise identification of drug targets and reducing the time and costs associated with developing new medications, particularly for complex diseases[56][57]. Developed by Google DeepMind and Isomorphic Labs, AlphaFold 3 builds upon the success of its predecessor, AlphaFold 2, by providing accurate atomic-level views of the structure of biomolecular systems. This includes not only proteins but also DNA, RNA, and small molecule ligands, along with their interactions and structural impacts due to post-translational modifications and ions[11][55]. }} \\
\texttt{\small{ The AI model's ability to predict complex protein interactions and structures with high accuracy offers a new set of drug target candidates to explore, potentially leading to groundbreaking therapeutic developments[56][57]. Furthermore, the application of AlphaFold 3 in predicting the structural impact of various molecular systems opens up exciting possibilities for rational drug development against targets that were previously difficult to modulate[55]. }} \\
\texttt{\small{ Although the initial impact of AlphaFold and similar models like RoseTTAFold on drug discovery has been incremental, the potential commercial and scientific value of AlphaFold 3 is vast, with its transformative potential already being acknowledged as "Nobel Prize-worthy"[19][21]. By accurately predicting the three-dimensional shapes of proteins and other biomolecules, AlphaFold 3 helps streamline the process of identifying compounds that will successfully bind to these targets, producing beneficial health outcomes[57]. }} \\
\texttt{\small{ Moreover, integrating AlphaFold 3 with emerging technologies such as self-driving laboratories could further accelerate and innovate the drug discovery process. The combination of AlphaFold 3's structural predictions with automated, high-throughput experimentation could dramatically speed up the validation and optimization of new drug candidates, transforming our understanding and approach to drug R\&D[9][44][55]. }} \\
\vspace{1pt}
\texttt{\small{\textbf{ \#\# ProteinDNA and ProteinRNA Interaction Predictions }}} \\
\texttt{\small{ AlphaFold 3 represents a significant advancement in the prediction of biomolecular interactions, specifically those involving proteins, DNA, and RNA. Unlike its predecessor, AlphaFold 2, which primarily focused on predicting the structure of individual proteins, AlphaFold 3 employs a diffusion-based architecture to predict raw atom coordinates. This allows it to model a variety of biomolecular interactions with high accuracy, including those between proteins and nucleic acids such as DNA and RNA[29]. }} \\
\texttt{\small{ Introduced in 2024 by Google DeepMind and Isomorphic Labs, AlphaFold 3 expands its predictive capabilities beyond proteins to encompass all of life's molecules. This includes small molecules known as ligands, which are significant in the context of drug discovery[31]. The ability to predict interactions between proteins and DNA holds particular promise for advancing genetic regulation understanding and developing targeted gene therapies[29][31]. }} \\
\texttt{\small{ The predictive power of AlphaFold 3 extends to complex biomolecular interactions, including those involving protein complexes with DNA, RNA, and various ligands and ions. This enhanced capability allows for a more comprehensive understanding of biological processes and has the potential to unlock transformative scientific developments, from biorenewable materials to more resilient crops and accelerated genomics research[34]. Additionally, AlphaFold 3's success rate of approximately 70\% in accurately predicting protein-protein interactions underscores its effectiveness[34]. }} \\
\texttt{\small{ Perhaps one of the most exciting aspects of AlphaFold 3 is its ability to model interactions between proteins and a wide range of biological molecules, including DNA and RNA. This advancement is critical for understanding the fundamental mechanisms of life and for identifying potential drug candidates, reflecting the extensive training set that includes a broad spectrum of molecules[53]. By accurately predicting these complex interactions, AlphaFold 3 has the potential to revolutionize various fields within biological research and biotechnology. }} \\
\vspace{1pt}
\texttt{\small{\textbf{ \#\# Efficiency and Accuracy Improvements in Drug Discovery }}} \\
\texttt{\small{ AlphaFold 3 has significantly improved the efficiency and accuracy of drug discovery processes, enabling more precise identification of drug targets and reducing the time and costs associated with developing new medications. This advancement is particularly impactful in the context of complex diseases, where traditional methods have struggled to provide swift and accurate results[7][20][59]. }} \\
\texttt{\small{ One notable example is the discovery of a more potent hit molecule, ISM042-2-048, through AI-powered compound generation. This compound demonstrated good inhibitory activity against CDK20, a crucial protein in hepatocellular carcinoma (HCC), with an IC50 value of 33.4 ± 22.6 nM. It also showed selective anti-proliferation activity in an HCC cell line, marking the first instance of AlphaFold being applied to hit identification in drug discovery[22]. Furthermore, scientists at Insilico and the University of Toronto have integrated AlphaFold into an end-to-end AI-powered drug discovery platform, leading to the identification of a new drug for a novel target aimed at treating HCC[23]. }} \\
\texttt{\small{ AlphaFold 3 has also enhanced prediction accuracy for antibody-antigen interactions, a critical area for immunology and therapeutic antibody development. By blending bioinformatics and physics, AlphaFold offers a more precise understanding of the exact binding between antibodies and antigens, surpassing the capabilities of previous computational methods[24]. }} \\
\texttt{\small{ Moreover, AlphaFold opens new avenues for exploring drug targets, especially in neglected diseases. These are conditions that receive little research funding due to affecting small or low-income populations, making them less attractive to commercial markets. The expanded scope of AlphaFold 3 to include a diverse range of biomolecules further paves the way for transformational science, including bio-renewable materials and more resilient crops, alongside accelerating drug discovery and genomics research[25][59]. }} \\
\vspace{1pt}
\texttt{\small{\textbf{ \#\# Biotechnology Applications }}} \\
\texttt{\small{ In 2024, together with Isomorphic Labs, we introduced AlphaFold 3, which predicts the structure and interactions of all of life’s molecules[1]. AlphaFold 3 goes beyond proteins to a broad spectrum of biomolecules including DNA, RNA, and even small molecules, also known as ligands, which encompass many drugs[1]. }} \\
\vspace{1pt}
\texttt{\small{\textbf{ \#\# Genetic Regulation and Personalized Medicine }}} \\
\texttt{\small{ AlphaFold 3 has heralded a significant advancement in our understanding of genetic regulation and the development of personalized medicine. By leveraging a diffusion-based architecture, AlphaFold 3 can predict the structure and interactions of various biomolecular systems with unprecedented accuracy, including proteins, nucleic acids, small molecules, ions, and modified residues[8][35]. This enhanced capability allows for joint structure prediction of complex biological systems, which is critical for understanding the intricate interactions within cells[35]. }} \\
\texttt{\small{ One of the most groundbreaking features of AlphaFold 3 is its ability to predict protein-DNA and protein-RNA interactions with far greater accuracy compared to previous models. The new AlphaFold model has shown a significant improvement—up to 50\% or more—in predicting these interactions, which are essential for understanding gene regulation and expression[30]. Such precise predictions could lead to revolutionary discoveries in the mechanisms of gene regulation, potentially transforming genetic research and opening new avenues for personalized medicine[30][35]. }} \\
\texttt{\small{ Furthermore, AlphaFold 3's ability to model how DNA interacts with proteins offers profound insights into cellular processes and the regulation of genetic codes. This capability can significantly advance our understanding of genetic regulation and help in the development of targeted gene therapies[50][52]. The improved accuracy in predicting these molecular interactions means that scientists can now explore genetic pathways with a level of detail previously unattainable, facilitating the creation of more effective and personalized treatment plans for various genetic disorders[8][52]. }} \\
\texttt{\small{ Despite these advancements, it is important to note that access to AlphaFold 3 is currently restricted, which may limit the widespread application of its capabilities in the short term[51]. However, the potential implications for genetic research and personalized medicine remain vast and promising as the technology continues to evolve and become more accessible to the scientific community. }} \\
\vspace{1pt}
\texttt{\small{\textbf{ \#\# Economic Impact and Market Implications }}} \\
\texttt{\small{ AlphaFold 3, a groundbreaking artificial intelligence program developed by Google DeepMind and Isomorphic Labs, has been heralded for its transformative potential in drug discovery and development, which could have substantial economic implications[21]. The program predicts the structure and interactions of all of life's molecules with remarkable accuracy, a significant advancement in the field of genetics[32][67]. }} \\
\texttt{\small{ By dramatically reducing the cost and time associated with protein structure determination, AlphaFold 3 has the potential to expedite research and development processes in the pharmaceutical industry, leading to significant cost savings[66]. This reduction in costs can benefit multiple sectors, particularly companies focused on biotechnology and pharmaceuticals. Stocks of companies involved in these sectors, such as Alphabet Inc., ABBV, EXAI, IBM, MRK, MSFT, NVS, ORCL, PFE, SDGR, and SLP, could see a positive impact due to the advancements brought about by AlphaFold 3[68]. }} \\
\texttt{\small{ Moreover, the program's ability to predict protein structures has practical applications in addressing global health challenges. For instance, researchers at the University of Cambridge are utilizing AlphaFold to develop a more effective malaria vaccine, while teams at the University of Colorado are exploring solutions to antibiotic resistance, a major public health concern[70]. These efforts underscore the potential for AlphaFold 3 to contribute to significant medical breakthroughs, which could further bolster investor confidence in related sectors. }} \\
\texttt{\small{ However, it remains uncertain how exactly AlphaFold 3 will catalyze drug discovery and development. Many drugs fail to reach the market due to unforeseen interactions between their components and various parts of the body[71]. Despite this uncertainty, the AI-driven advancements of AlphaFold 3 hold promise for overcoming some of these traditional hurdles, thereby accelerating the pipeline from research to market. }} \\

\texttt{\small{\textbf{ \# References }}} \\

\texttt{\small\text{[1]} AlphaFold - Google DeepMind. Retrieved June 3, 2024, from \url{https://deepmind.google/technologies/alphafold/}} \\
\texttt{\small\text{[2]} AlphaFold - Wikipedia. Retrieved June 3, 2024, from \url{https://en.wikipedia.org/wiki/AlphaFold}} \\
\texttt{\small\text{[3]} AlphaFold 3 predicts the structure and interactions of all of life’s molecules - Isomorphic Labs. Retrieved June 3, 2024, from \url{https://www.isomorphiclabs.com/articles/alphafold-3-predicts-the-structure-and-interactions-of-all-of-lifes-molecules}} \\
\texttt{\small\text{[4]} AlphaFold 3 predicts the structure and interactions of all of life’s molecules. Retrieved June 3, 2024, from \url{https://blog.google/technology/ai/google-deepmind-isomorphic-alphafold-3-ai-model/}} \\
\texttt{\small\text{[5]} AlphaFold 3 upgrade enables the prediction of other types of biomolecular systems. Retrieved June 3, 2024, from \url{https://phys.org/news/2024-05-alphafold-enables-biomolecular.html}} \\
\texttt{\small\text{[6]} Rational drug design with AlphaFold 3 - Isomorphic Labs. Retrieved June 3, 2024, from \url{https://www.isomorphiclabs.com/articles/rational-drug-design-with-alphafold-3}} \\
\texttt{\small\text{[7]} Google DeepMind’s AlphaFold 3 Could Transform Drug Discovery. Retrieved June 3, 2024, from \url{https://time.com/6975934/google-deepmind-alphafold-3-ai/}} \\
\texttt{\small\text{[8]} Accurate structure prediction of biomolecular interactions with AlphaFold 3 | Nature. Retrieved June 3, 2024, from \url{https://www.nature.com/articles/s41586-024-07487-w}} \\
\texttt{\small\text{[9]} AlphaFold 3 predicts the structure and interactions of all of life’s molecules. Retrieved June 3, 2024, from \url{https://blog.google/technology/ai/google-deepmind-isomorphic-alphafold-3-ai-model/}} \\
\texttt{\small\text{[10]} A glimpse of the next generation of AlphaFold - Google DeepMind. Retrieved June 3, 2024, from \url{https://deepmind.google/discover/blog/a-glimpse-of-the-next-generation-of-alphafold/}} \\
\texttt{\small\text{[11]} Rational drug design with AlphaFold 3 - Isomorphic Labs. Retrieved June 3, 2024, from \url{https://www.isomorphiclabs.com/articles/rational-drug-design-with-alphafold-3}} \\
\texttt{\small\text{[12]} Accurate structure prediction of biomolecular interactions with AlphaFold 3 | Nature. Retrieved June 3, 2024, from \url{https://www.nature.com/articles/s41586-024-07487-w}} \\
\texttt{\small\text{[13]} AlphaFold - Google DeepMind. Retrieved June 3, 2024, from \url{https://deepmind.google/technologies/alphafold/}} \\
\texttt{\small\text{[14]} AlphaFold 3: Google DeepMind’s latest AI tech in drug discovery. Retrieved June 3, 2024, from \url{https://www.prescouter.com/2024/05/google-deepmind-alphafold-3/}} \\
\texttt{\small\text{[15]} AlphaFold - Wikipedia. Retrieved June 3, 2024, from \url{https://en.wikipedia.org/wiki/AlphaFold}} \\
\texttt{\small\text{[16]} AlphaFold 3 upgrade enables the prediction of other types of biomolecular systems. Retrieved June 3, 2024, from \url{https://phys.org/news/2024-05-alphafold-enables-biomolecular.html}} \\
\texttt{\small\text{[17]} AlphaFold 3: A Leap Forward in Biomolecular Structure Prediction — Opportunities and Limitations | by Freedom Preetham | Meta Multiomics | May, 2024 | Medium. Retrieved June 3, 2024, from \url{https://medium.com/meta-multiomics/alphafold-3-a-leap-forward-in-biomolecular-structure-prediction-opportunities-and-limitations-924350af1181}} \\
\texttt{\small\text{[18]} AlphaFold3 and its improvements in comparison to AlphaFold2 | by Falk Hoffmann | May, 2024 | Medium. Retrieved June 3, 2024, from \url{https://medium.com/@falk_hoffmann/alphafold3-and-its-improvements-in-comparison-to-alphafold2-96815ffbb044}} \\
\texttt{\small\text{[19]} What does AlphaFold mean for drug discovery?. Retrieved June 3, 2024, from \url{https://www.nature.com/articles/d41573-021-00161-0}} \\
\texttt{\small\text{[20]} DeepMind's latest AlphaFold model is more useful for drug discovery | TechCrunch. Retrieved June 3, 2024, from \url{https://techcrunch.com/2023/10/31/deepminds-latest-alphafold-model-is-more-useful-for-drug-discovery/}} \\
\texttt{\small\text{[21]} Why AlphaFold 3 is stirring up so much buzz in pharma | PharmaVoice. Retrieved June 3, 2024, from \url{https://www.pharmavoice.com/news/google-alphafold-3-drug-discovery-pharma-buzz/716496/}} \\
\texttt{\small\text{[22]} AlphaFold accelerates artificial intelligence powered drug discovery: efficient discovery of a novel CDK20 small molecule inhibitor - PMC. Retrieved June 3, 2024, from \url{https://www.ncbi.nlm.nih.gov/pmc/articles/PMC9906638/}} \\
\texttt{\small\text{[23]} First Application of AlphaFold in Identifying Potential Liver Cancer Drug. Retrieved June 3, 2024, from \url{https://www.genengnews.com/topics/drug-discovery/first-application-of-alphafold-in-identifying-potential-liver-cancer-drug/}} \\
\texttt{\small\text{[24]} AlphaFold 3: Google DeepMind’s latest AI tech in drug discovery. Retrieved June 3, 2024, from \url{https://www.prescouter.com/2024/05/google-deepmind-alphafold-3/}} \\
\texttt{\small\text{[25]} AlphaFold Is The Most Important Achievement In AI—Ever. Retrieved June 3, 2024, from \url{https://www.forbes.com/sites/robtoews/2021/10/03/alphafold-is-the-most-important-achievement-in-ai-ever/}} \\
\texttt{\small\text{[26]} AlphaFold 3 predicts the structure and interactions of all of life’s molecules. Retrieved June 3, 2024, from \url{https://blog.google/technology/ai/google-deepmind-isomorphic-alphafold-3-ai-model/}} \\
\texttt{\small\text{[27]} AlphaFold 3 offers even more accurate protein structure prediction. Retrieved June 3, 2024, from \url{https://www.drugdiscoverytrends.com/meet-alphafold-3-which-can-accurately-model-more-than-99-of-molecular-types-in-the-protein-data-bank/}} \\
\texttt{\small\text{[28]} A glimpse of the next generation of AlphaFold - Google DeepMind. Retrieved June 3, 2024, from \url{https://deepmind.google/discover/blog/a-glimpse-of-the-next-generation-of-alphafold/}} \\
\texttt{\small\text{[29]} AlphaFold 3 offers even more accurate protein structure prediction. Retrieved June 3, 2024, from \url{https://www.drugdiscoverytrends.com/meet-alphafold-3-which-can-accurately-model-more-than-99-of-molecular-types-in-the-protein-data-bank/}} \\
\texttt{\small\text{[30]} AlphaFold 3 predicts the structure and interactions of all of life’s molecules - Isomorphic Labs. Retrieved June 3, 2024, from \url{https://www.isomorphiclabs.com/articles/alphafold-3-predicts-the-structure-and-interactions-of-all-of-lifes-molecules}} \\
\texttt{\small\text{[31]} AlphaFold - Google DeepMind. Retrieved June 3, 2024, from \url{https://deepmind.google/technologies/alphafold/}} \\
\texttt{\small\text{[32]} AlphaFold 3 predicts the structure and interactions of all of life’s molecules - Isomorphic Labs. Retrieved June 3, 2024, from \url{https://www.isomorphiclabs.com/articles/alphafold-3-predicts-the-structure-and-interactions-of-all-of-lifes-molecules}} \\
\texttt{\small\text{[33]} AlphaFold 3 predicts the structure and interactions of all of life’s molecules. Retrieved June 3, 2024, from \url{https://blog.google/technology/ai/google-deepmind-isomorphic-alphafold-3-ai-model/}} \\
\texttt{\small\text{[34]} AlphaFold - Wikipedia. Retrieved June 3, 2024, from \url{https://en.wikipedia.org/wiki/AlphaFold}} \\
\texttt{\small\text{[35]} AlphaFold 3 upgrade enables the prediction of other types of biomolecular systems. Retrieved June 3, 2024, from \url{https://phys.org/news/2024-05-alphafold-enables-biomolecular.html}} \\
\texttt{\small\text{[36]} AlphaFold 3 offers even more accurate protein structure prediction. Retrieved June 3, 2024, from \url{https://www.drugdiscoverytrends.com/meet-alphafold-3-which-can-accurately-model-more-than-99-of-molecular-types-in-the-protein-data-bank/}} \\
\texttt{\small\text{[37]} A glimpse of the next generation of AlphaFold - Google DeepMind. Retrieved June 3, 2024, from \url{https://deepmind.google/discover/blog/a-glimpse-of-the-next-generation-of-alphafold/}} \\
\texttt{\small\text{[38]} AlphaFold3: A Game Changer in Protein Structure Prediction — Part 1 | by Chithra Srinivasan | May, 2024 | Medium. Retrieved June 3, 2024, from \url{https://medium.com/@csn289/alphafold3-a-game-changer-in-protein-structure-prediction-part-1-b8d9c361bda2}} \\
\texttt{\small\text{[39]} AlphaFold 3 offers even more accurate protein structure prediction. Retrieved June 3, 2024, from \url{https://www.drugdiscoverytrends.com/meet-alphafold-3-which-can-accurately-model-more-than-99-of-molecular-types-in-the-protein-data-bank/}} \\
\texttt{\small\text{[40]} Frontiers | Before and after AlphaFold2: An overview of protein structure prediction. Retrieved June 3, 2024, from \url{https://www.frontiersin.org/articles/10.3389/fbinf.2023.1120370/full}} \\
\texttt{\small\text{[41]} AlphaFold 3: A Leap Forward in Biomolecular Structure Prediction — Opportunities and Limitations | by Freedom Preetham | Meta Multiomics | May, 2024 | Medium. Retrieved June 3, 2024, from \url{https://medium.com/meta-multiomics/alphafold-3-a-leap-forward-in-biomolecular-structure-prediction-opportunities-and-limitations-924350af1181}} \\
\texttt{\small\text{[42]} AlphaFold - Wikipedia. Retrieved June 3, 2024, from \url{https://en.wikipedia.org/wiki/AlphaFold}} \\
\texttt{\small\text{[43]} Rational drug design with AlphaFold 3 - Isomorphic Labs. Retrieved June 3, 2024, from \url{https://www.isomorphiclabs.com/articles/rational-drug-design-with-alphafold-3}} \\
\texttt{\small\text{[44]} The Drug Discoverer - Reflecting on DeepMind’s AlphaFold artificial intelligence success – what’s the real significance for protein folding research and drug discovery? - The Institute of Cancer Research, London. Retrieved June 3, 2024, from \url{https://www.icr.ac.uk/blogs/the-drug-discoverer/page-details/reflecting-on-deepmind-s-alphafold-artificial-intelligence-success-what-s-the-real-significance-for-protein-folding-research-and-drug-discovery}} \\
\texttt{\small\text{[45]} AlphaFold Is The Most Important Achievement In AI—Ever. Retrieved June 3, 2024, from \url{https://www.forbes.com/sites/robtoews/2021/10/03/alphafold-is-the-most-important-achievement-in-ai-ever/}} \\
\texttt{\small\text{[46]} AlphaFold 3 predicts the structure and interactions of all of life’s molecules - Isomorphic Labs. Retrieved June 3, 2024, from \url{https://www.isomorphiclabs.com/articles/alphafold-3-predicts-the-structure-and-interactions-of-all-of-lifes-molecules}} \\
\texttt{\small\text{[47]} Great expectations – the potential impacts of AlphaFold DB | EMBL. Retrieved June 3, 2024, from \url{https://www.embl.org/news/science/alphafold-potential-impacts/}} \\
\texttt{\small\text{[48]} AlphaFold: Accelerating biological research - Google DeepMind. Retrieved June 3, 2024, from \url{https://deepmind.google/impact/meet-the-scientists-using-alphafold/}} \\
\texttt{\small\text{[49]} How is AlphaFold2 used by scientists? | AlphaFold. Retrieved June 3, 2024, from \url{https://www.ebi.ac.uk/training/online/courses/alphafold/validation-and-impact/how-is-alphafold-used-by-scientists/}} \\
\texttt{\small\text{[50]} AlphaFold 3 Will Change the Biological World and Drug Discovery. Retrieved June 3, 2024, from \url{https://www.analyticsvidhya.com/blog/2024/05/deepmind-ai-alphafold/}} \\
\texttt{\small\text{[51]} Major AlphaFold upgrade offers boost for drug discovery. Retrieved June 3, 2024, from \url{https://www.nature.com/articles/d41586-024-01383-z}} \\
\texttt{\small\text{[52]} AlphaFold 3 upgrade enables the prediction of other types of biomolecular systems. Retrieved June 3, 2024, from \url{https://phys.org/news/2024-05-alphafold-enables-biomolecular.html}} \\
\texttt{\small\text{[53]} AlphaFold 3: Stepping into the future of structure prediction - Front Line Genomics. Retrieved June 3, 2024, from \url{https://frontlinegenomics.com/alphafold-3-stepping-into-the-future-of-structure-prediction/}} \\
\texttt{\small\text{[54]} Analyzing the potential of AlphaFold in drug discovery | MIT News | Massachusetts Institute of Technology. Retrieved June 3, 2024, from \url{https://news.mit.edu/2022/alphafold-potential-protein-drug-0906}} \\
\texttt{\small\text{[55]} Rational drug design with AlphaFold 3 - Isomorphic Labs. Retrieved June 3, 2024, from \url{https://www.isomorphiclabs.com/articles/rational-drug-design-with-alphafold-3}} \\
\texttt{\small\text{[56]} AlphaFold 3: Google DeepMind’s latest AI tech in drug discovery. Retrieved June 3, 2024, from \url{https://www.prescouter.com/2024/05/google-deepmind-alphafold-3/}} \\
\texttt{\small\text{[57]} AlphaFold Is The Most Important Achievement In AI—Ever. Retrieved June 3, 2024, from \url{https://www.forbes.com/sites/robtoews/2021/10/03/alphafold-is-the-most-important-achievement-in-ai-ever/}} \\
\texttt{\small\text{[58]} AlphaFold 3: Revolutionizing drug discovery and molecular biology. Retrieved June 3, 2024, from \url{https://www.prescouter.com/2024/05/alphafold-3/}} \\
\texttt{\small\text{[59]} Google DeepMind’s AI model AlphaFold 3 can be a gamechanger in drug discovery. Retrieved June 3, 2024, from \url{https://indiaai.gov.in/article/google-deepmind-s-ai-model-alphafold-3-can-be-a-gamechanger-in-drug-discovery}} \\
\texttt{\small\text{[60]} AlphaFold Is The Most Important Achievement In AI—Ever. Retrieved June 3, 2024, from \url{https://www.forbes.com/sites/robtoews/2021/10/03/alphafold-is-the-most-important-achievement-in-ai-ever/}} \\
\texttt{\small\text{[61]} New study uses AlphaFold and AI to accelerate design of novel drug for liver cancer. Retrieved June 3, 2024, from \url{https://acceleration.utoronto.ca/news/new-study-uses-alphafold-and-ai-to-accelerate-design-of-novel-drug-for-liver-cancer}} \\
\texttt{\small\text{[62]} AlphaFold 3: Google DeepMind’s latest AI tech in drug discovery. Retrieved June 3, 2024, from \url{https://www.prescouter.com/2024/05/google-deepmind-alphafold-3/}} \\
\texttt{\small\text{[63]} AlphaFold Is The Most Important Achievement In AI—Ever. Retrieved June 3, 2024, from \url{https://www.forbes.com/sites/robtoews/2021/10/03/alphafold-is-the-most-important-achievement-in-ai-ever/}} \\
\texttt{\small\text{[64]} The rise of self-driving labs in chemical and materials sciences | Nature Synthesis. Retrieved June 3, 2024, from \url{https://www.nature.com/articles/s44160-022-00231-0}} \\
\texttt{\small\text{[65]} Self-driving laboratories to autonomously navigate the protein fitness landscape | Nature Chemical Engineering. Retrieved June 3, 2024, from \url{https://www.nature.com/articles/s44286-023-00002-4}} \\
\texttt{\small\text{[66]} r/stocks on Reddit: Which stocks will benefit most from alphafold protein fold prediction advancement?. Retrieved June 3, 2024, from \url{https://www.reddit.com/r/stocks/comments/k7z8hi/which_stocks_will_benefit_most_from_alphafold/}} \\
\texttt{\small\text{[67]} Did Google's DeepMind Just Revolutionize Medicine? Retrieved June 3, 2024, from \url{https://www.fool.com/investing/2020/12/05/did-googles-deepmind-just-revolutionize-medicine/}} \\
\texttt{\small\text{[68]} Alphabet’s AlphaFold Is A Gift To Humanity, Not Investors | Seeking Alpha. Retrieved June 3, 2024, from \url{https://seekingalpha.com/article/4522954-alphabets-alphafold-is-a-gift-to-humanity-not-investors}} \\
\texttt{\small\text{[69]} DeepMind's AlphaFold AI Origin Story, Influence on Biotech. Retrieved June 3, 2024, from \url{https://www.businessinsider.com/deepmind-alphafold-ai-origin-story-impact-on-biotech-2023-2}} \\
\texttt{\small\text{[70]} AlphaFold - Google DeepMind. Retrieved June 3, 2024, from \url{https://deepmind.google/technologies/alphafold/}} \\
\texttt{\small\text{[71]} Folds and faults: On AlphaFold 3 - The Hindu. Retrieved June 3, 2024, from \url{https://www.thehindu.com/opinion/editorial/folds-and-faults-on-alphafold-3/article68161671.ece}} \\
\texttt{\small\text{[72]} Software Engineers Lead Drug Breakthough Terry 11:07 AM - Markman on Tech. Retrieved June 3, 2024, from \url{https://www.thestreet.com/tech/news/sdgrjdm121420}} \\
\bottomrule
\caption{\system's generated final report for ``Alpha Fold 3''. ``\#'', ``\#\#'' indicate the section title and subsection title respectively. Numbers in brackets indicate the cited references.}
\label{table:example_final_report}
\end{longtable}

\end{document}